\documentclass[journal]{IEEEtran}
\ifCLASSINFOpdf
\else
\fi
\usepackage{amssymb}
\usepackage{amsthm}
\usepackage{lineno}
\usepackage{graphicx}
\usepackage[caption=false,font=footnotesize]{subfig}
\usepackage{amsmath}
\usepackage{float}
\usepackage{hyperref}
\usepackage{pifont}
\usepackage{makecell}
\usepackage{array}
\usepackage{booktabs}
\usepackage[figuresright]{rotating}
\usepackage{multirow}
\usepackage{makecell}
\usepackage{url}
\usepackage{subcaption}

\usepackage{bbding}
\usepackage{wasysym}
\usepackage{subcaption}
\usepackage{soul}
\usepackage{color, xcolor}
\usepackage{cite}
\soulregister\cite7
\soulregister\eqref7
\soulregister\ref7
\definecolor{change}{RGB}{255,251,80}

\begin{document}
%
\title{Structure-guided Diffusion Transformer for Low-Light Image Enhancement}
%
%
%

\author{
    Xiangchen Yin, Zhenda Yu, Longtao Jiang, Xin Gao, Xiao Sun, Senior Member, IEEE, Zhi Liu, Senior Member, IEEE, Xun Yang, Member, IEEE
    \thanks{Xiangchen Yin, Longtao Jiang, and Xun Yang are associated with the University of Science and Technology of China. Xiangchen Yin is also with the Institute of Artificial Intelligence, Hefei Comprehensive National Science Center, Hefei 230088, China. Email: {\tt\small yinxiangchen@mail.ustc.edu.cn}, {\tt\small taotao707@mail.ustc.edu.cn}, {\tt\small xyang21@ustc.edu.cn}}

   \thanks{Zhenda Yu is associated with Anhui University, China, and also with the Institute of Artificial Intelligence, Hefei Comprehensive National Science Center, Hefei 230088, China. Email:  {\tt\small wa22201140@stu.ahu.edu.cn}}%
    \thanks{Xin Gao is associated with the School of Vehicle and Mobility, Tsinghua University, China, and also with the State Key Laboratory of Automotive Safety and Energy, Tsinghua University, Beijing, China. Email: {\tt\small gaoxin97@mail.tsinghua.edu.cn}}

    \thanks{Xiao Sun is associated with the School of Computer Science and Information Engineering, Hefei University of Technology, China, and also with the Institute of Artificial Intelligence, Hefei Comprehensive National Science Center, Hefei 230088, China. Email: {\tt\small sunx@hfut.edu.cn}}
    
    \thanks{
        Zhi Liu is associated with Department of Computer and Network Engineering, The University of Electro-Communications, Chofu-shi, Tokyo,1828585 Japan. Email: {\tt\small liu@ieee.org}
    }
    \thanks{
    This work was supported by Special Project of the National Natural Science Foundation of China (62441614), Anhui Province Key R\&D Program (202304a05020068) and General Programmer of the National Natural Science Foundation of China (62376084). This research was also supported by the advanced computing resources provided by the Supercomputing Center of the USTC. We also acknowledge the support of GPU cluster built by MCC Lab of Information Science and Technology Institution, USTC. (Corresponding authors: Xiao Sun and Xun Yang.)
    }

}

\markboth{IEEE Transactions on Multimedia,~Vol.~14, No.~8, August~2015}%
{Shell \MakeLowercase{\textit{et al.}}: Bare Demo of IEEEtran.cls for IEEE Journals}
%



\maketitle

\begin{abstract}
While the diffusion transformer (DiT) has become a focal point of interest in recent years, its application in low-light image enhancement remains a blank area for exploration. Current methods recover the details from low-light images while inevitably amplifying the noise in images, resulting in poor visual quality. In this paper, we firstly introduce DiT into the low-light enhancement task and design a novel Structure-guided Diffusion Transformer based Low-light image enhancement (SDTL) framework. We compress the feature through wavelet transform to improve the inference efficiency of the model and capture the multi-directional frequency band. Then we propose a Structure Enhancement Module (SEM) that uses structural prior to enhance the texture and leverages an adaptive fusion strategy to achieve more accurate enhancement effect. In Addition, we propose a Structure-guided Attention Block (SAB) to pay more attention to texture-riched tokens and avoid interference from noisy areas in noise prediction. Extensive qualitative and quantitative experiments demonstrate that our method achieves SOTA performance on several popular datasets, validating the effectiveness of SDTL in improving image quality and the potential of DiT in low-light enhancement tasks. 

\end{abstract}


%
\IEEEpeerreviewmaketitle

\section{Introduction}

Low-light images bring unpleasant visibility and result in complex detail damage (e.g., noise, color distortion, etc.). Therefore, low-light enhancement  has received increasing attention in computer vision to improve the visual quality of images and support downstream vision applications \cite{loh2019getting, yin2023pe, wang2024eulermormer, wang2024frequency}.

Traditional enhancement methods are mainly based on image priors or physical models. Histogram equalization \cite{pizer1990contrast} adjusts the pixel distribution of the image through statistics. Retinex-based methods \cite{land1977retinex, fu2016weighted} decompose the image into illumination components and reflection components and adjust the reflection components. However, these methods often generate unnatural images and have poor scene robustness due to the use of artificially designed priors. Deep learning-based methods \cite{Chen2018Retinex, jiang2021enlightengan,zhang2024adaptive,hao2020low,ma2022retinex} fit complex mappings from low-light to normal-light on paired datasets. However, these methods have difficulty in recovering weak and missing details, even amplifying the noise to a certain extent inevitably.

\begin{figure}
\centerline{\includegraphics[width=1.0\linewidth]{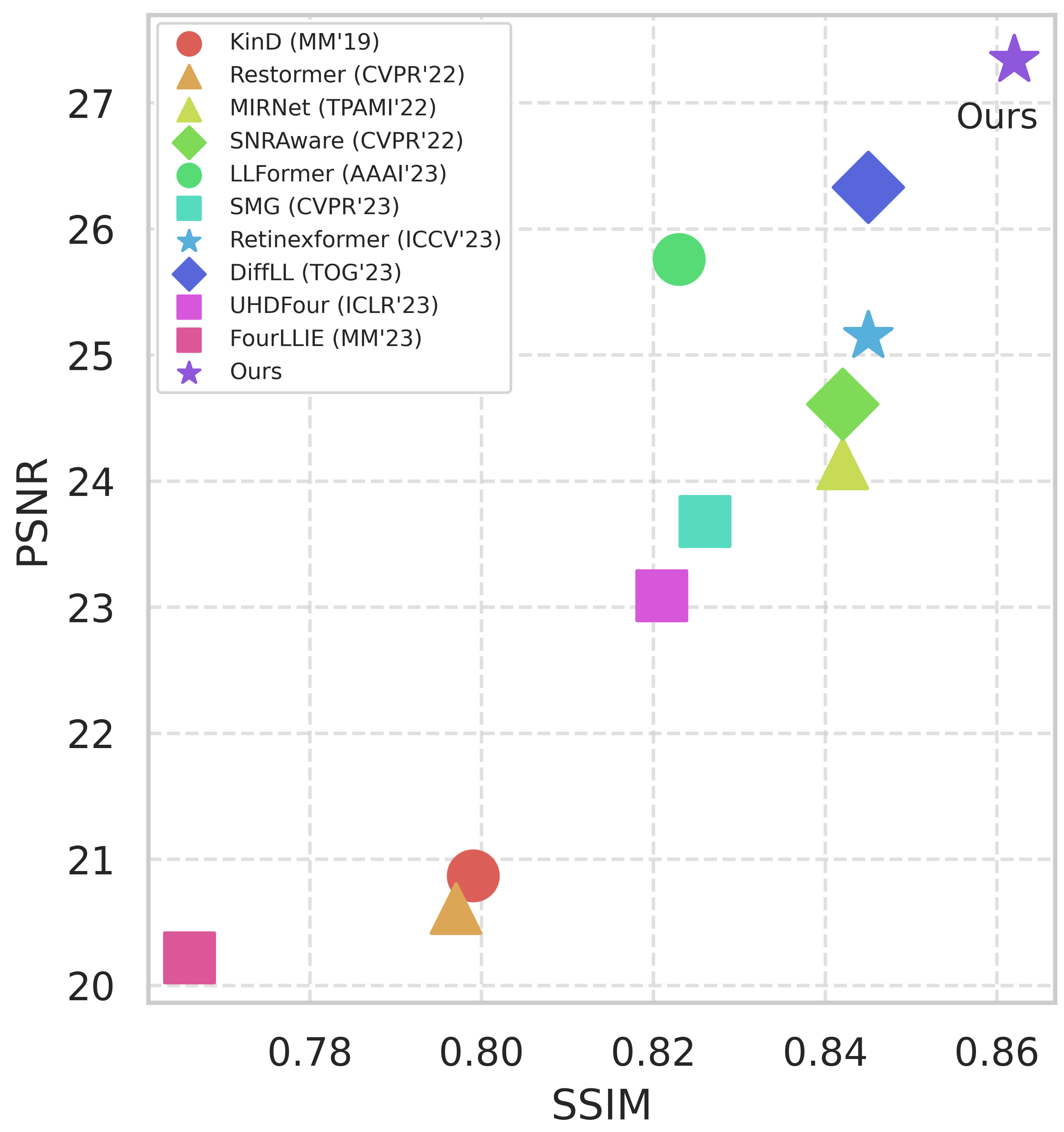}}
\vspace{-0.3cm}
\caption{Performance comparison between our method and other SOTA methods on the LOLv1 \cite{Chen2018Retinex} dataset. Our method has achieved the best results on both PSNR and SSIM, reaching 27.34 and 0.862 respectively. The legend on the right shows each model and source. } 
\vspace{-0.3cm}
\label{Vis_Com}
\end{figure}

Recently the denoising diffusion probabilistic model (DDPM) \cite{ho2020denoising} has demonstrated remarkable success in the domain of image generation, which produces more realistic details through a series of refinement. Recognizing the prowess of DDPM in accurately capturing pixel distributions, it has been increasingly leveraged for enhancing image quality. In addition, there have been several successful attempts to introduce conditional DDPM in low-level vision tasks \cite{jiang2023low, saharia2022image, saharia2022palette}. However, the diffusion model requires longer training time and the image recovery effect is unstable. Transformer \cite{vaswani2017attention} obtains long-range dependencies in sequences and captures the global context in images. The diffusion model stands at the cutting edge of image generation, offering substantial benefits. However, the prevalent reliance on the rudimentary U-Net architecture fails to fully capitalize on the extensive potential of the diffusion model, suggesting a need for more sophisticated structural adaptations. Therefore, the research on the diffusion transformer (DiT) \cite{peebles2023scalable} for low-light enhancement needs to be deeply explored. The structure of DiT not only keeps the advantages of the transformer but also freely controls the size of the model by adjusting parameters such as the number of blocks and the embedded dimension.

\begin{figure*}
\centerline{\includegraphics[width=1.0\linewidth]{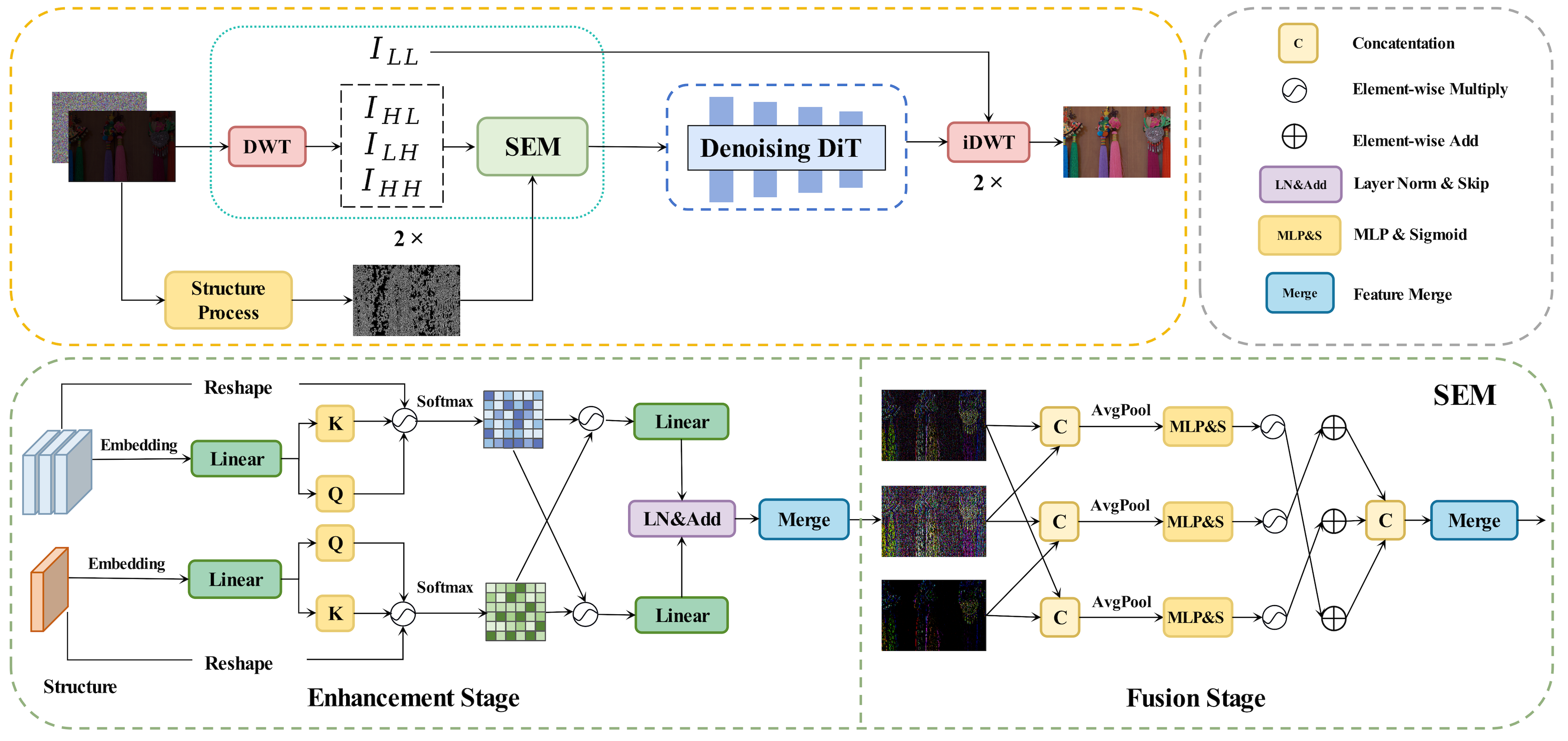}}
\caption{Overview of SDTL. Our model compresses features to improve the efficiency of inference through wavelet transform. We design a Structure Enhancement Module (SEM) to enhance structural information under different frequency bands, which is divided into two stages: \textbf{Enhancement} and \textbf{Fusion}. In addition, we propose a Structure-guided Attention Block (SAB) to pay attention to the texture-riched tokens in the noise prediction network.} \label{Overview}
\vspace{-0.3cm}
\end{figure*}

To solve the above problems, this paper introduces the DiT technique for the low-light enhancement task for the first time and achieve impressive performance. We propose a novel \textbf{S}tructure-guided \textbf{D}iffusion \textbf{T}ransformer based \textbf{L}ow-light (SDTL) enhancement framework, making full use of generation capabilities of the diffusion model. Firstly, we transform the features into the frequency domain via wavelet transformation to improve the inference efficiency of the model, and the features are decomposed into the information in multiple directions, including horizontal, vertical, and diagonal. This decomposition captures local structures at different scales, providing richer information for subsequent enhancements. Then we design a Structure Enhancement Module (\textbf{SEM}) to enhance the frequency domain information in two steps: \textit{enhancement} and \textit{fusion}. SEM enhances the ability of network to refine details by incorporating structural priors, effectively directing the focus of network towards critical structural elements and simultaneously muting the influence of extraneous details. Additionally, we design an adaptive fusion strategy to complement the information in various directions. As for DiT, we propose a Structure-guided Attention Block (\textbf{SAB}) to guide the self-attention mechanism between tokens. We notice that only local areas of the image corresponding to texture-rich tokens are beneficial to the network, so we adopt structural priors to pay more attention to these areas and avoid influence from noisy areas. We conducted extensive experiments on 8 benchmark datasets, and the results show that our SDTL surpasses the current state-of-the-art (SOTA) models in low-light enhancement tasks without adopting any training strategy, as shown in Fig. \ref{Vis_Com}. Experimental results demonstrate the effectiveness of our method in high image quality and model efficiency, while also reflect the great potential of DiT on low-light enhancement.

Our contributions can be summarized below:
\begin{itemize}

\item We 
devise a novel \textbf{S}tructure-guided \textbf{D}iffusion \textbf{T}rans-former based \textbf{L}ow-light image enhancement (\textbf{SDTL}) method, which achieves impressive performance.

\item We design a Structure Enhancement Module (\textbf{SEM}) to exploit contextual information. By introducing structural priors, SEM enables the network to focus on important structural cues while suppressing other irrelevant information, thus improving the visual quality of the enhanced image. In addition, we also use an adaptive fusion strategy to complement information in various directions.

\item We propose a Structure-guided Attention Block (\textbf{SAB}) in DiT to guide the self-attention mechanism between tokens so that the network pays more attention to texture-rich areas and avoids interference from noise areas.

\item Extensive experiments on multiple datasets show that our SDTL surpasses current state-of-the-art models and improves the performance of night segmentation.

\end{itemize}

\section{Related Work}

\subsection{Low-light Image Enhancement}
Previous work used some traditional techniques to enhance images such as histogram equalization \cite{pizer1990contrast}, gamma correction \cite{huang2012efficient}, and Retinex adjustment \cite{guo2016lime, fu2016weighted}. With the prosperity of deep learning, many learning-based methods \cite{moran2020deeplpf, lim2020dslr,wang2023pmsnet,wu2023cycle} have been proposed to learn the mapping from low-light to normal directly through CNN or Transformer. Wei et al. \cite{Chen2018Retinex} proposed RetinexNet, which combines Retinex theory and CNN to achieve learnable image decomposition and adjustment of the reflection component. Xu et al. \cite{xu2022snr} jointed signal-noise ratio (SNR) and Transformer to achieve spatially varying image enhancement. LLFormer \cite{wang2023ultra} was proposed to enhance ultra-high definition (UHD) images, and they published a UHD low-light dataset. Wang et al. \cite{wang2023fourllie} proposed FourLLIE, which estimates the amplitude transform in Fourier space to improve the brightness of low-light images.

\subsection{Diffusion for Low-level Vision}
Recently, with the advancement of the denoising diffusion probabilistic model (DDPM), great progress has been made in the diffusion model for low-level vision \cite{gao2023implicit,sahak2023denoising,nguyen2024diffusion}, which makes it gradually used in low-level vision tasks. Xia et al. \cite{xia2023diffir} proposed an efficient DiffIR, which guides the convergence of the diffusion model through a compact IR prior representation. Jiang et al. \cite{jiang2023low} used the diffusion model to perform image enhancement in the wavelet domain, which significantly speeds up the inference speed, but ignores the prior knowledge of enhancement. Whang et al. \cite{whang2022deblurring} proposed a deblurring framework based on the conditional diffusion model, which trained a random sampler to refine the output of the predictor. Yin et al \cite{yin2023cle} proposed the Controllable Light Enhancement Diffusion model (CLE Diffusion), which allows users to input the required brightness level and apply the SAM model to achieve friendly interactive regional controllable brightness. However, the performance gap between this method and previous SOTA methods is not large and there is no great improvement in quantitative indicators. The previous method adopts U-Net shape denoising network, the flexibility of the model parameters is limited and the sampling speed is slow. Our method improves the speed through wavelet transform compression features and uses diffusion transformer to set network parameters at will according to the actual situation.

\section{Methods}

Given an image $I$, we select a pair of low-pass/high-pass filters to perform 2-dim Discrete Wavelet Transform (DWT) for frequency band decomposition as:
\begin{equation}
I_\textrm{LL}, I_\textrm{HL}, I_\textrm{LH}, I_\textrm{HH} = \mathcal{W}[I; G_\textrm{L}(t), G_\textrm{H}(t)],
\end{equation}
where $\mathcal{W}$ represents the 2-dim DWT, $G_\textrm{L}$ represents a low-pass filter, $G_\textrm{H}$ represents a high-pass filter, $I_\textrm{LL}$, $I_\textrm{HL}$, $I_\textrm{LH}$, $I_\textrm{HH}$ represents low-frequency subband, vertical high-frequency subband, horizontal high-frequency subband, and diagonal high-frequency subband respectively. The low-frequency subband is home to the overarching structural essence of the image, capturing its fundamental layout and composition. In contrast, the high-frequency subband is where the intricate texture details reside, often accompanied by noise elements that are particularly prevalent in low-light images. We design a Structure Enhancement Module (SEM) to enhance structural information of high-frequency. After two wavelet transforms, we all adopted a SEM and the features will be reduced to 16 times of the original. Note that the low-frequency information doesn't participate in the network, and is only used for the final Inverse Wavelet Transform (IWT). We take the features as a condition after frequency processing and concatenate it with random noise as the input of the diffusion model. The overview of our SDTL is shown in Fig. \ref{Overview}.

\subsection{Conditional Diffusion Model}

The diffusion model gradually converts the distribution of Gaussian noise into a data distribution by learning the Markov chain, which is divided into two stages: forward diffusion and reverse denoising. Gaussian noise is gradually added to the forward diffusion. Firstly, we define a variance scheduler with $T$ diffusion steps $\{\beta_{1}, \beta_{2}, \beta_{3}, \ldots, \beta_{T}\}$:
\begin{equation}
q\left(\mathbf{x}_{1: T} \mid \mathbf{x}_{0}\right)=\prod_{t=1}^{T} \mathcal{N}\left(\mathbf{x}_{t} ; \sqrt{1-\beta_{t}} \mathbf{x}_{t-1}, \beta_{t} \mathbf{I}\right),
\end{equation}
where $x_t$ is the damaged data after noising, $\mathcal{N}$ denotes the Gaussian noise, $x_0$ denotes the original image, and $I$ denotes the condition of the diffusion model. The data of time $t$ is only related to time $t-1$. The larger the $t$, the closer it is to pure noise. The reverse denoising process gradually denoises the randomly sampled Gaussian noise, and the model will predict the reverse distribution of noise $p_{\theta}$ as:
\begin{equation}
p_{\theta}\left(X_{0: T}\right)  =p\left(x_{T}\right) \prod_{t=1}^{T} p_{\theta}\left(x_{t-1} \mid x_{t}\right), 
\end{equation}
\begin{equation}
p_{\theta}\left(x_{t-1} \mid x_{t}\right)  =\mathcal{N}\left(x_{t-1} ; \mu_{\theta}\left(x_{t}, t\right), \sigma_t^2\mathcal{I}\right),
\end{equation}
where $\boldsymbol{\epsilon}_{\theta}$ denotes a noise prediction network, $\theta$ is the parameters of $\boldsymbol{\epsilon}_{\theta}$ (here is DiT), $\mu_{\theta}\left(x_{t}, t\right)$ and $\sigma_t^2$ are the mean and variance of the distribution respectively. We adopt the frequency of low-light images as a condition, so the variance is constrained by $\mathcal{I}$, and the mean is computed as:
\begin{align}
\boldsymbol{\mu}_{\theta}\left(\mathbf{x}_{t}, t\right) &= \frac{1}{\sqrt{\alpha_{t}}}\left(\mathbf{x}_{t}-\frac{\beta_{t}}{\sqrt{1-\bar{\alpha}_{t}}} \boldsymbol{\epsilon}_{\theta}\left(\mathbf{x}_{t}, t\right)\right), \\
\alpha_t &= 1-\beta_t, \bar{\alpha}_{t}=\prod_{i=1}^{T} \alpha_{i}.
\end{align}
The optimization goal of diffusion model  is to maximize the logarithmic likelihood of the data distribution, and the loss function is defined as
\begin{equation}
\mathcal{L}=E_{\mathbf{x}_{0}, t, \epsilon_{t} \sim \mathcal{N}(\mathbf{0}, \mathbf{I})}\left[\left\|\epsilon_{t}-\epsilon_{\theta}\left(\mathbf{x}_{t}, t\right)\right\|^{2}\right]. 
\end{equation}

\subsection{Structure Enhancement Module}
Give three frequency bands in different directions after the wavelet transform $\mathcal{I}_\textrm{HL}$, $\mathcal{I}_\textrm{LH}$, $\mathcal{I}_\textrm{HH} \in R^{H \times W \times C}$ and the structural information initially processed by convolution $\mathcal{S} \in R^{H \times W \times C}$. We design a Structure Enhancement Module (SEM) to achieve guided enhancement and multi-directional complementary fusion, as shown in Fig. \ref{Overview}. Then we introduce the two stages of enhancement and fusion of SEM.

\noindent{\textbf{(1) Enhancement}}: It should be noted that we perform the same operation for each frequency band. We firstly flatten the features and structure of a single frequency band to $R^{N \times C} (N=H \times W)$. Then two linear layers are used to transform the features as two vectors $\mathcal{X}_{I}$ and $\mathcal{X}_{S}$ ($R^{N \times \hat{C}}$), respectively. We further adopt a novel bi-directional guidance mechanism, which spreads the information of vectors with each other to make up for the uncertainty of structural information on the dark area, and provides more efficient structural guidance to the frequency band. Then we encode $\mathcal{X}_{I}$ and $\mathcal{X}_{S}$ into vectors respectively, which get the key $\mathcal{K}_{I}$ and query $\mathcal{Q}_{I}$ of the features, and the key $\mathcal{K}_{S}$ and query $\mathcal{Q}_{S}$ of the structure. We multiply the attention map with another sequence to realize the exchange of information between the sequence and the sequence. This process is described as:
\begin{align}
\mathcal{E}_I &= \textrm{Softmax}(\frac{\mathcal{K}_{I}^{T} \cdot \mathcal{Q}_{I}} {\alpha}) \cdot \mathcal{R}(\mathcal{X}_{S}), \\
\mathcal{E}_S &= \textrm{Softmax}(\frac{\mathcal{K}_{S}^{T} \cdot \mathcal{Q}_{S}} {\alpha}) \cdot \mathcal{R}(\mathcal{X}_{I}), 
\end{align}
where $\mathcal{R}$ denotes the reshape of the vector and $\alpha$ denotes the scale factor. After the information interaction, the vectors undergo processing via a linear layer, and then we implement layer normalization and skip connection. Finally, we perform the merge operation through a convolutional block and reshape the feature back to $R^{H \times W \times C}$ to get the enhanced frequency $\hat{\mathcal{I}_t} (t=\textrm{HL}, \textrm{LH}, \textrm{HH})$.

\begin{figure}[tbp]
  \centering
    \begin{minipage}{0.49\linewidth}
    \includegraphics[width=\linewidth]{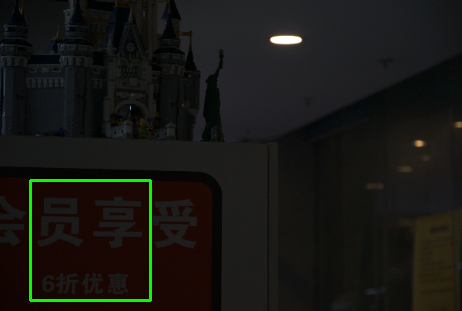}
    \centerline{Origin image}
  \end{minipage}
  \hfill
    \begin{minipage}{0.49\linewidth}
    \includegraphics[width=\linewidth]{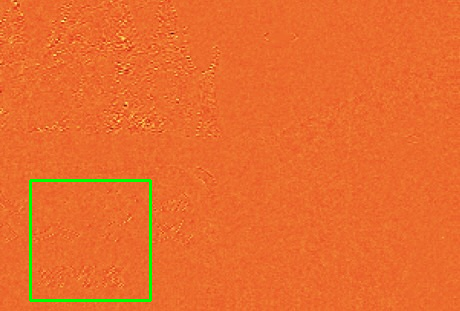}
    \centerline{Initial feature}
  \end{minipage}
    \\[1.0mm]
    \begin{minipage}{0.49\linewidth}
    \includegraphics[width=\linewidth]{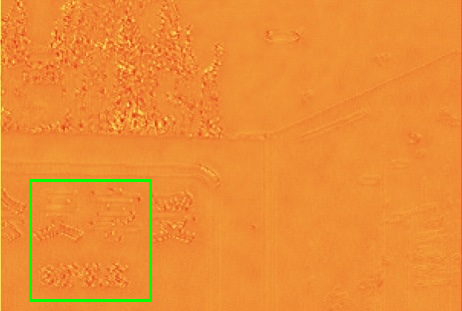}
    \centerline{Enhanced feature}
  \end{minipage}
  \hfill
    \begin{minipage}{0.49\linewidth}
    \includegraphics[width=\linewidth]{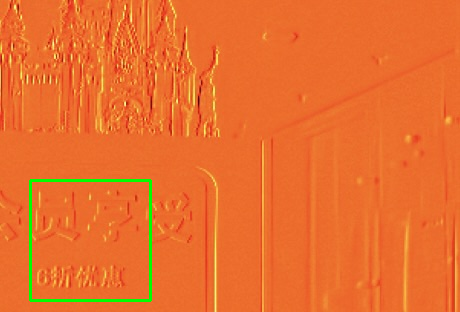}
    \centerline{Fused feature}
  \end{minipage}
    \\
    
  \caption{Visualization of hot feature for SEM. We visualize the initial feature, enhanced feature, and fused feature in SEM respectively. It can be found that our SEM has a significant enhancement effect on features.}\label{Feature_vis}
\end{figure}

\noindent{\textbf{(2) Fusion}}: The frequency bands in different directions after enhancement are usually complementary. In this stage, we propose a new cross-band fusion mechanism, and each band is complementarily integrated after being corrected by the other two bands. Firstly, after the current frequency band is processed with convolution blocks, the other two frequency bands are concat on the channel, and the features are adjusted by the channel gate, then cross into the current frequency band by addition operation:
\begin{equation}
\begin{aligned}
&\hat{\mathcal{I}{\delta}} = \hat{\mathcal{I}{\delta}} + \mathcal{G}(\textrm{Concat}( \hat{\mathcal{I}{\kappa}}, \hat{\mathcal{I}{\tau}} )), \\
&\delta,\kappa,\tau \in \{\textrm{HL}, \textrm{LH}, \textrm{HH}\}; \delta \neq \kappa \neq \tau 
\end{aligned}
\end{equation}
where $\mathcal{G}(x)=\sigma( \textrm{MLP}(\textrm{AvgPool}(x)) ) * x$ represents channel gate, $\sigma$ represents sigmoid function and $\textrm{AvgPool}$ represents average pooling. Finally, we use a convolution block to summarize the information of the three frequency bands and make a skip connection. We visualize the hot features of the enhancement and fusion stages in SEM, as shown in Fig. \ref{Feature_vis}. We can find the change of text in the rectangle part, by visualizing origin image, initial feature, enhanced feature and fused feature for comparison at different stages. Comparing the "Initial feature" and "Enhanced feature", shows that the structural prior significantly restores high-frequency information in the enhancement stage for subsequent noise prediction. Comparing "Enhanced feature" and "Fused feature" shows that fusion makes the frequency bands in different directions complementary and more balanced in color and texture.


\subsection{Denoising of Diffusion Transformer}
\begin{figure}
\centerline{\includegraphics[width=1.0\linewidth]{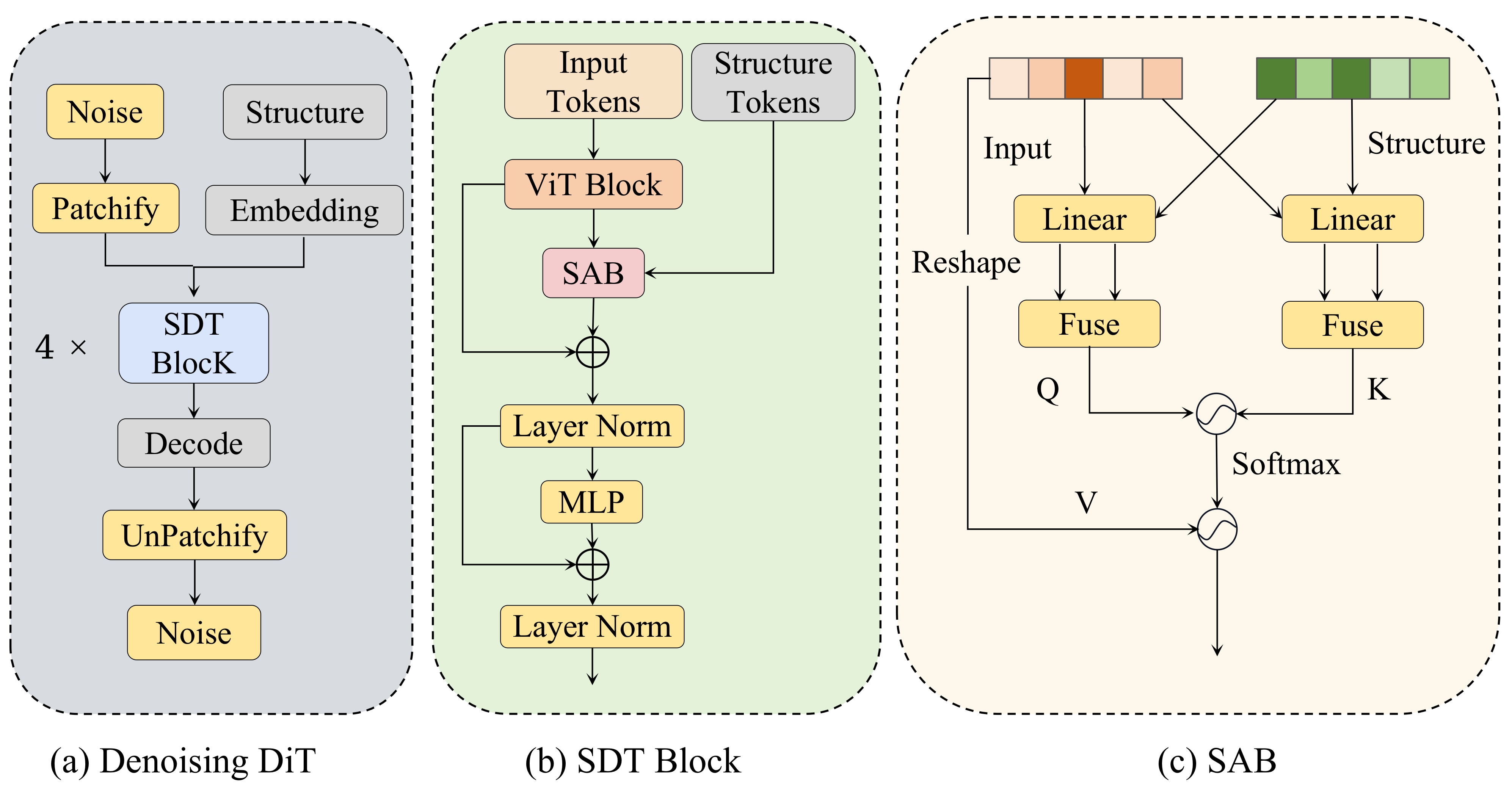}}
\caption{Details of Denoising Diffusion Transformer (a). SDT block consists of a ViT block and a SAB block (b). SAB integrates the original token with the structural token to make the network focus on the textured-riched token (c). } \label{SDT}
\vspace{-0.3cm}
\end{figure}

In this section, we introduce the structural design of DiT, as shown in Fig. \ref{SDT}. Give conditional input $C \in R^{H \times W \times 6}$, we convert the feature diagram of spatial input into $N$ token sequences through patchify, $N$ is determined by the size of each patch $p$, it is calculated as $N=HW/p^2$, the embedding dimension of each token is set to 384. In this paper, we set $p$ to 4 and the head number of self-attention is set to 6.
In addition, we adopt 6 Structure DiT (SDT) blocks, each of which mainly consists of a ViT \cite{dosovitskiy2020image} block and a Structure-guided Attention Block (SAB). SAB (Fig. \ref{SDT} (c)) introduces structural information to guide the self-attention mechanism between tokens, which pays more attention to texture-riched token sequences and avoids influence by noisy areas. After patchify and projection the structure sequence is fused into the token sequence of features for self-attention calculation. This process is expressed as
\begin{equation}
\begin{aligned}
&Q,K=\textrm{Fuse}(\mathcal{P}(x) , \mathcal{P}(\mathcal{S})),\\
&V=\mathcal{R}(\mathcal{P}(x)),\\
&\textrm{SAB}(x) = \textrm{Softmax}(K^TQ/\alpha) V.
\end{aligned}
\end{equation}
where $\mathcal{P}$ denotes the patchify and $\textrm{Fuse}$ denotes the cross attention operation that fuses the two inputs. We use 4 SDT blocks as the encoding layer and use skip connection. Finally, the decoding layer uses two SDT blocks and a linear projection layer, and then rearranges the token to restore the same size as the original feature $R^{H \times W \times 3}$ to get the predicted noise.

\begin{table*}
  \caption{Quantitative comparisons of different methods on the LOLv1 \cite{Chen2018Retinex}, LOLv2 \cite{yang2021sparse} and LSRW \cite{hai2023r2rnet} datasets. The best results and second-best results are highlighted in blue and green, respectively. Note that we obtained these results either from the corresponding papers, or by running the officially released models, and the missing results on LSRW are marked as ``-''.}\label{comparisons-main}
  \renewcommand{\arraystretch}{1.2} 

  \centering
  \setlength{\tabcolsep}{1.5mm}{
  \begin{tabular}{l|c|cccc|cccc|cccc}
    \toprule[1.2pt]
    \multirow{2}{*}{\textbf{Methods}} &\multirow{2}{*}{\textbf{References}}  &\multicolumn{4}{|c|}{\textbf{LSRW}} &\multicolumn{4}{c}{\textbf{LOLv1}}  &\multicolumn{4}{|c}{\textbf{LOLv2-real}} \\ 
    \cline{3-14}
    & &PSNR$\uparrow$  &SSIM$\uparrow$  &LPIPS$\downarrow$ &FID$\downarrow$
	&PSNR$\uparrow$  &SSIM$\uparrow$  &LPIPS$\downarrow$ &FID$\downarrow$
	&PSNR$\uparrow$  &SSIM$\uparrow$  &LPIPS$\downarrow$ &FID$\downarrow$ \\
   \hline
    RetinexNet \cite{Chen2018Retinex} &BMVC'18  &15.61 &0.414 &0.454 & 108.350 &16.77 &0.462 &0.474  &126.266  &18.37 &0.723 &0.365 & 133.905 \\ 
    KinD \cite{zhang2019kindling} &MM'19  &- &- &- &- &20.87 &0.799 &0.207  &104.632  &17.54 &0.669 &0.375 &137.346 \\ 
    DRBN \cite{yang2020fidelity} &CVPR'20  &16.73 &0.507 &0.457 &80.727 &19.86 &0.834 &0.155  & 98.732  &20.13 &0.830 &0.147 & 89.085 \\ 
    EnlightenGAN \cite{jiang2021enlightengan} &TIP'21   &17.11 &0.463 &0.406 &69.033 &17.48 &0.652 &0.322  &94.704  &18.64 &0.677 &0.309  & 84.044\\ 
    Restormer \cite{zamir2022restormer} &CVPR'22  &16.30 &0.453 &0.427 &69.219 &20.61 &0.797 &0.288 &72.998 &24.91 &0.851 &0.264 &58.649 \\ 
    URetinex-Net \cite{wu2022uretinex} &CVPR'22  &18.27 &0.518 &0.419 &66.871&19.84 &0.824 &0.237 &52.383 &21.09 &0.858 &0.208 &49.836 \\ 
    Uformer \cite{wang2022uformer} &CVPR'22  &16.59 &0.494 &0.435 &82.299 &19.00 &0.741 &0.354 &109.351 &18.44 &0.759 &0.347 &98.138 \\ 
    MIRNet \cite{zamir2022learning} &TPAMI'22  &16.47 &0.477 &0.430 &93.811 &24.14 &0.842 &0.131 & 69.179 &20.36 &0.782 &0.317 & 49.108\\ 
    SNR-Aware \cite{xu2022snr} &CVPR'22  &16.49 &0.505 &0.419 & 65.807  &24.61 &0.842 &0.152   & 55.121 &21.48 &0.849 &0.157 & 54.532 \\ 
    IAT \cite{cui2022illumination} &BMVC'22  &\textcolor{green}{20.81} &0.565 &0.467 &80.499 &23.38 &0.809 &0.134 &67.412 &23.50 &0.824 &0.191 & 62.153 \\
    LLFormer \cite{wang2023ultra} &AAAI'23 &20.69 &0.560 &0.518 &96.782 &25.76 &0.823 &0.167  &65.271  &26.20 &0.819 &0.209 &63.432 \\ 
    FourLLIE \cite{wang2023fourllie} &MM'23 &18.61 &0.505 &\textcolor{green}{0.316} &73.550 &20.22 &0.766 &0.250 &91.793 &22.34 &0.847 &\textcolor{blue}{0.051} & 89.334\\ 
    UHDFour \cite{li2023embedding} &ICLR'23 &17.30 &0.529 &0.443 &62.032 &23.09 &0.821 &0.259 &56.912 &21.79 &0.854 &0.292 &60.837 \\ 
    SMG \cite{xu2023low} &CVPR'23 &19.04 &\textcolor{green}{0.568} &0.392 &101.560 &23.68 &0.826 &\textcolor{green}{0.118} &58.846 &24.62 &0.867 &0.148 & 78.582  \\ 
    Retinexformer \cite{cai2023retinexformer} &ICCV'23 &20.15 &0.534 &0.336 &70.360 &25.15 &0.845 &0.131 &71.148 &22.80 &0.840 &0.171 &62.439 \\ 
    DiffLL \cite{jiang2023low} &TOG'23 &19.28 &0.552 &0.350 & \textcolor{green}{45.294} &\textcolor{green}{26.33} &0.845 &0.217 &\textcolor{blue}{48.114} &\textcolor{green}{28.85} &\textcolor{green}{0.876} &0.207 &\textcolor{green}{45.359}  \\ 
    RSFNet \cite{saini2024specularity} &CVPR'24 &- &- &- & - &22.15 &0.860 &0.265 &- &21.59 &0.843 &0.278 &-  \\
    LightenDiffusion \cite{jiang2025lightendiffusion} &ECCV'24 &18.55 &0.539 &\textcolor{blue}{0.311} & - &20.45 &0.803 &0.192 &- &- &- &- &-  \\
    ExpoMamba~\cite{adhikarla2024expomamba} &ICML'24 &- &- &- & - &25.77 &\textcolor{green}{0.860} &0.212 &89.210 &28.04 &\textcolor{blue}{0.885} &0.232 &85.920  \\
    
    \hline
    SDTL (Ours) &- &\textcolor{blue}{21.23} &\textcolor{blue}{0.576} &0.392 &\textcolor{blue}{42.569}  &\textcolor{blue}{27.34}	&\textcolor{blue}{0.862}	&\textcolor{blue}{0.118} &\textcolor{green}{50.432} &\textcolor{blue}{28.85}	&0.875	&\textcolor{green}{0.131}  &\textcolor{blue}{43.237} \\   
    \bottomrule[1.2pt]
  \end{tabular}}
\end{table*}

\section{EXPERIMENTS}

\noindent{\textbf{Datasets}}. We validate the effectiveness of our methods on the LOLv1 \cite{Chen2018Retinex}, LOLv2-real \cite{yang2021sparse}, LSRW \cite{hai2023r2rnet} datasets. The LOLv1 dataset has 485 paired images for training and 15 paired images for testing, both in sizes (400, 600). The LOLv2 dataset consists of a subset of real and synthetic scenes. LOLv2-real has 689 images for training and 100 images for testing respectively. LSRW is a large low-light dataset. It uses Huawei and Nikon cameras to collect 5,650 samples of different scenes, of which 5,600 are for training and 50 for testing. In addition, to verify the robustness of our model, we evaluate it on five datasets without reference: LIME \cite{guo2016lime}, VV \cite{vonikakis2008fast}, DICM \cite{lee2013contrast}, NPE \cite{wang2013naturalness}, and MEF \cite{ma2015perceptual}. The ACDC \cite{sakaridis2021acdc} dataset can be used to train semantic segmentation under adverse conditions. It contains 4,006 images, distributed in the four common bad weather of fog, night, rain, and snow, and each image has a high-quality pixel-level label. In this paper, we only use the part of the night scene to verify the performance of downstream night segmentation.

\begin{table}[t]
\caption{NIQE score on the LIME \cite{guo2016lime}, VV \cite{vonikakis2008fast}, DICM \cite{lee2013contrast}, NPE \cite{wang2013naturalness}, and MEF \cite{ma2015perceptual} datasets. The best results and second-best results are marked in blue and green respectively.}\label{NIQE_Compa}
\label{ablation}
\centering
\renewcommand{\arraystretch}{1.2} 
\begin{tabular}{l|ccccc|c}
\hline
\textbf{Methods}  & \textbf{LIME} & \textbf{VV} & \textbf{DICM} & \textbf{NPE} & \textbf{MEF} & \textbf{Average} \\
\hline
Zero-DCE \cite{guo2020zero} & 5.820 & 4.810 & 4.580 & 4.530 & 4.930 & 4.934\\
RetinexNet\cite{Chen2018Retinex} & 5.750 & 4.320 & 4.330 & 4.950 & 4.930 & 4.856\\
KinD\cite{zhang2019kindling}& 4.772  &  3.835 &  3.614 & 4.175 & 4.819 & 4.194 \\
MIRNet\cite{zamir2022learning} & 6.453  &  4.735 &  4.042 & 5.235 & 5.504 & 5.101	\\ 
SMG\cite{xu2023low}& 5.451  &  4.884  &  4.733 & 5.208 & 5.754 & 5.279  \\
FECNet\cite{huang2022deep} & 6.041  &  3.346  &  4.139 & 4.500 & 4.707 & 4.336	\\
FourLLIE\cite{wang2023fourllie} & 4.402 & 3.168 &\textcolor{blue}{3.374} & 3.909 & 4.362 & 3.907 \\
LLFormer\cite{wang2023ultra} & 4.796 & 5.030 & 4.080 & \textcolor{green}{3.809} & \textcolor{blue}{3.647} & 4.272 \\
DiffLL\cite{jiang2023low} & \textcolor{green}{4.203} & \textcolor{green}{3.081} & 3.953 & \textcolor{blue}{3.535} & 4.309 & \textcolor{green}{3.816} \\
\hline
STDL (Ours)  &\textcolor{blue}{4.004}  &\textcolor{blue}{2.927}  & \textcolor{green}{3.552} &3.977  &\textcolor{green}{3.729}
&\textcolor{blue}{3.565}\\ 
\hline
\end{tabular}
\end{table}

\noindent{\textbf{Evaluation Metrics}}. For the LOLv1, LOLv2 and LSRW datasets, we adopt the four metrics of PSNR, SSIM \cite{wang2004image}, LPIPS \cite{zhang2018unreasonable} and FID \cite{heusel2017gans} respectively to evaluate models. For the LIME, VV, DICM, NPE, and MEF datasets, we adopt the non-reference image quality metrics NIQE \cite{mittal2012making}, BRISQUE \cite{mittal2012no} and PI \cite{blau20182018} to evaluate them. On the nighttime segmentation dataset ACDC, we adopt $Precision$, $Recall$, $Fscore$ and $mIoU$ to test the performance of the model. $mIoU$ is the mean of $IoU$ in all classes, $IoU$ is described as
\begin{equation}
IoU= \frac{TP}{TP+FP+FN},  
\end{equation}
where $TP$ and $FP$ are true positive and false positive respectively, $FN$ is false negative. $Fscore$ takes into account $Precision$ and $Recall$, which is described as:
\begin{equation}
Fscore=\frac{2 \cdot \text { Precision } \cdot \text { Recall }}{\text { Precision } + \text { Recall }}, 
\end{equation} 
where $Precision=\frac{TP}{TP+FP}$ and $Recall=\frac{TP}{TP+FN}$.

\begin{table}[t]
\caption{{Application of night segmentation on the ACDC \cite{sakaridis2021acdc} dataset. We used DeepLabv3+ \cite{chen2018encoder} as the segmentor (Baseline) to enhance the images. The best results are marked in blue.} }\label{Segmentation}
\renewcommand{\arraystretch}{1.2} 
\centering
\begin{tabular}{l|cccc}
\hline
\textbf{Methods}  & \textbf{Precision}$\uparrow$ & \textbf{Recall}$\uparrow$ & \textbf{Fscore}$\uparrow$ & \textbf{mIoU}$\uparrow$  \\
\hline
Baseline &77.73	&69.29	&72.41	&59.57 \\
DeepLPF \cite{moran2020deeplpf}	&79.19	&71.31	&73.92	&61.88 \\
IAT \cite{cui2022you}	&79.02	&70.67	&73.64	&61.54 \\
FourLLIE \cite{wang2023fourllie}	&78.61	&69.79 	&72.84	&60.44 \\
DiffLL \cite{jiang2023low}	&78.30	&71.34	&73.78	&61.49 \\
SDTL (Ours)	& \textcolor{blue}{79.53}	& \textcolor{blue}{71.34}	&\textcolor{blue}{74.18}	 &\textcolor{blue}{62.25} \\
\hline
\end{tabular}
\end{table}

\begin{figure*}[t]
\begin{center}
\renewcommand\arraystretch{0.1}
\begin{tabular}{ccccccc}
\hspace{-4mm}
\includegraphics[width = 0.16\linewidth]{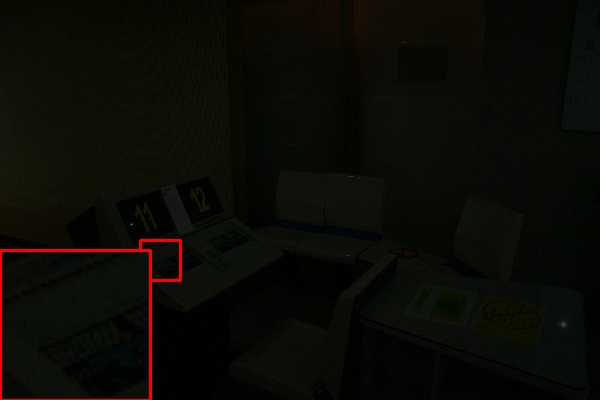} &\hspace{-4mm}
\includegraphics[width = 0.16\linewidth]{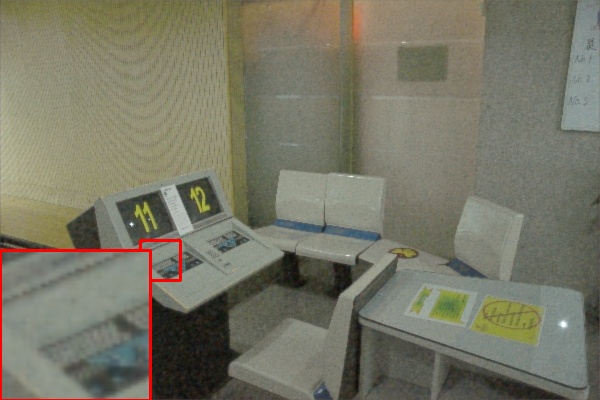} &\hspace{-4mm}
\includegraphics[width = 0.16\linewidth]{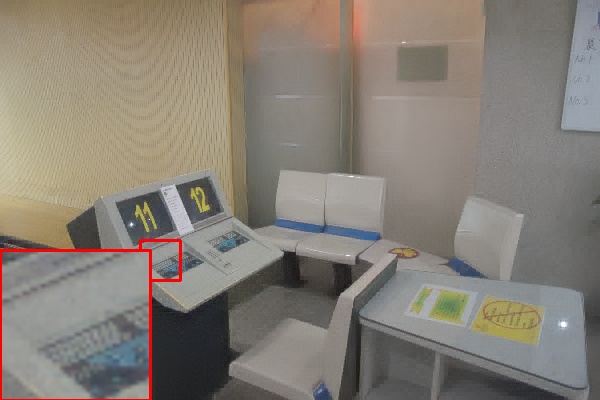} &\hspace{-4mm}
\includegraphics[width = 0.16\linewidth]{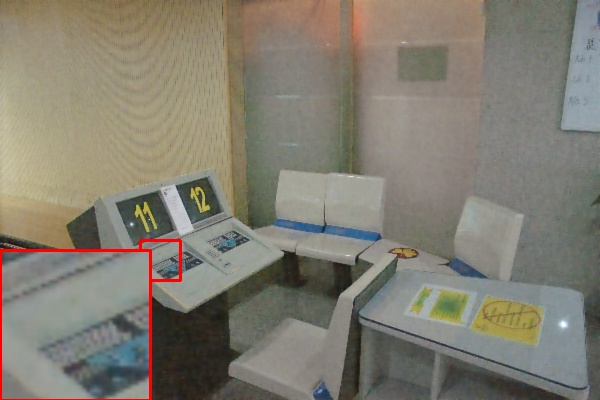} 
&\hspace{-4mm}
\includegraphics[width = 0.16\linewidth]{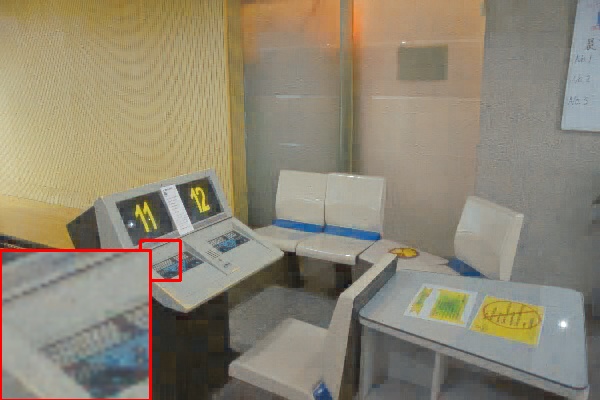} &\hspace{-4mm}
\includegraphics[width = 0.16\linewidth]{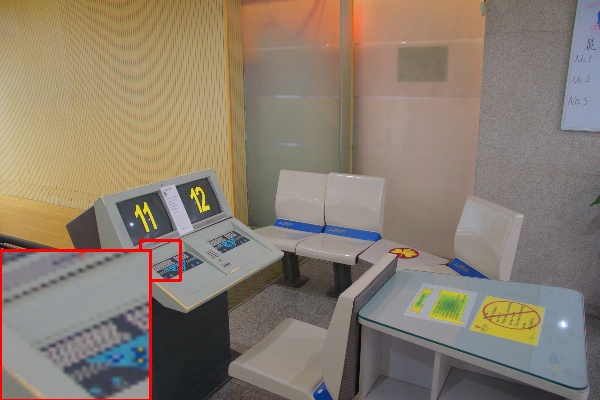}  &\hspace{-4mm}
\\[0.2mm]

\hspace{-4mm}
\includegraphics[width = 0.16\linewidth]{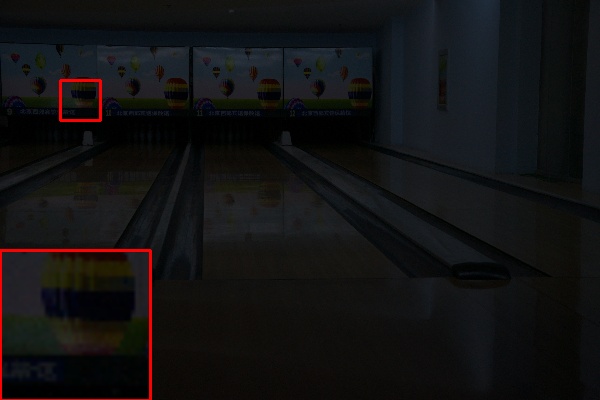} &\hspace{-4mm}
\includegraphics[width = 0.16\linewidth]{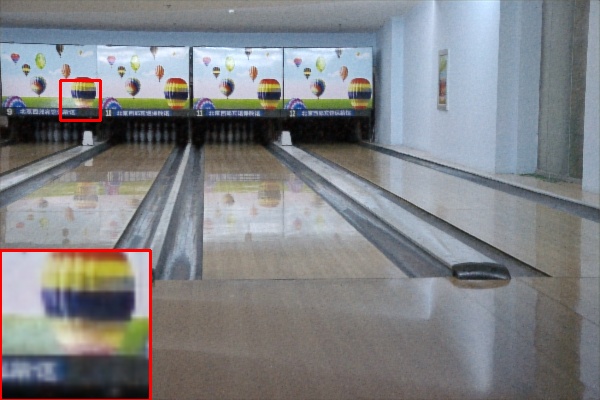} &\hspace{-4mm}
\includegraphics[width = 0.16\linewidth]{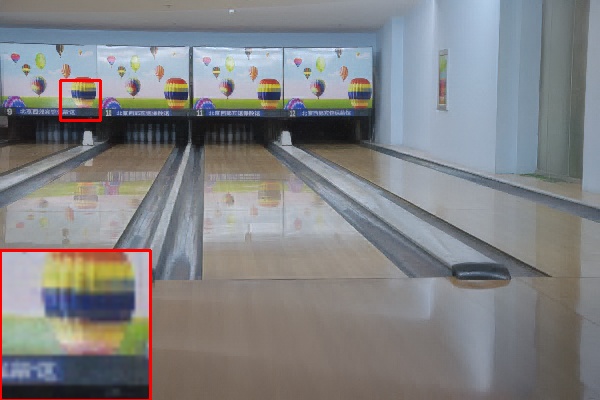} &\hspace{-4mm}
\includegraphics[width = 0.16\linewidth]{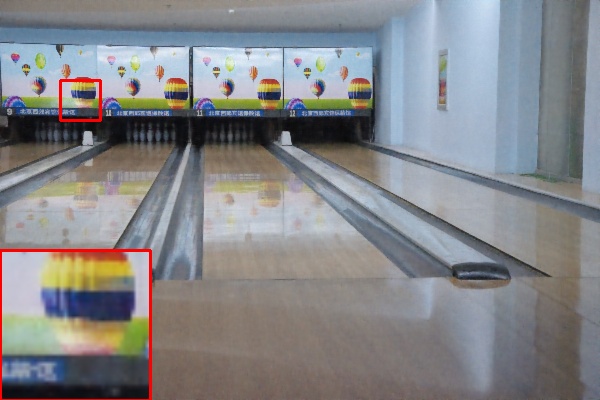} 
&\hspace{-4mm}
\includegraphics[width = 0.16\linewidth]{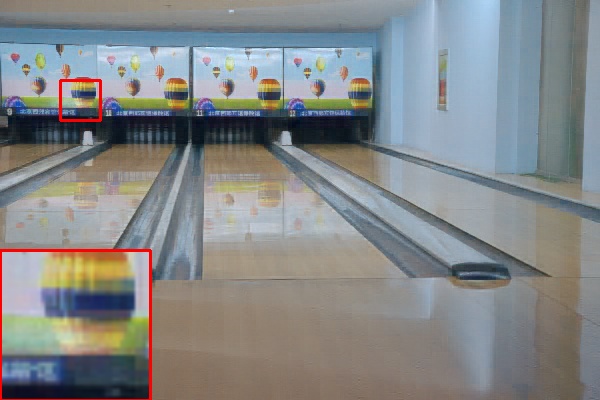} &\hspace{-4mm}
\includegraphics[width = 0.16\linewidth]{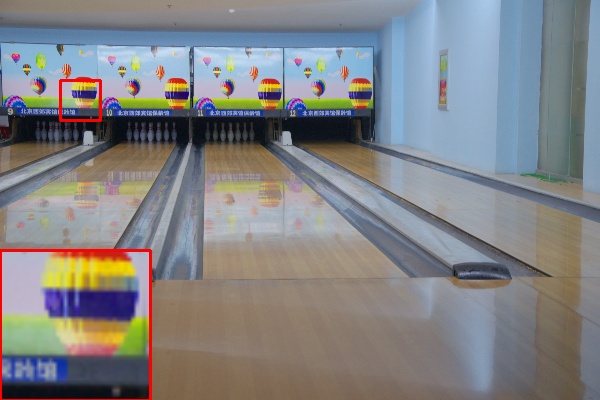} &\hspace{-4mm}
\\[0.2mm]

Input &\hspace{-4mm}    IAT &\hspace{-4mm} DiffLL  &\hspace{-4mm} Retinexformer &\hspace{-4mm} Ours &\hspace{-4mm}  GT 
\\[0.6mm]
\\

\hspace{-4mm}
\includegraphics[width = 0.16\linewidth]{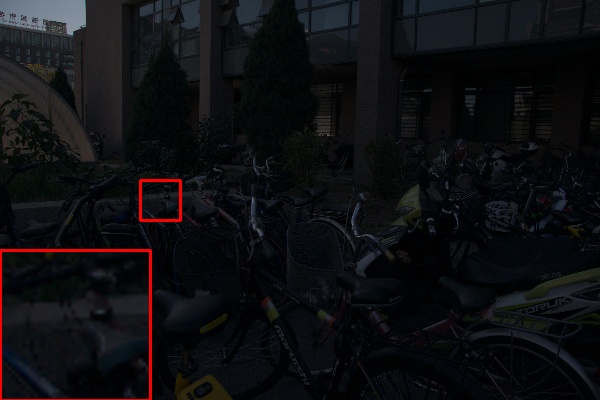} &\hspace{-4mm}
\includegraphics[width = 0.16\linewidth]{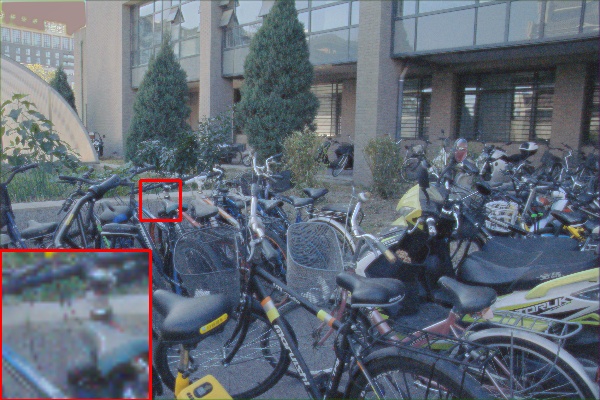} &\hspace{-4mm}

\includegraphics[width = 0.16\linewidth]{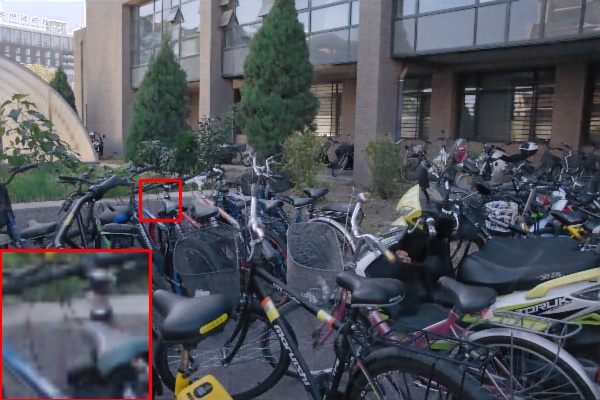} 
&\hspace{-4mm}
\includegraphics[width = 0.16\linewidth]{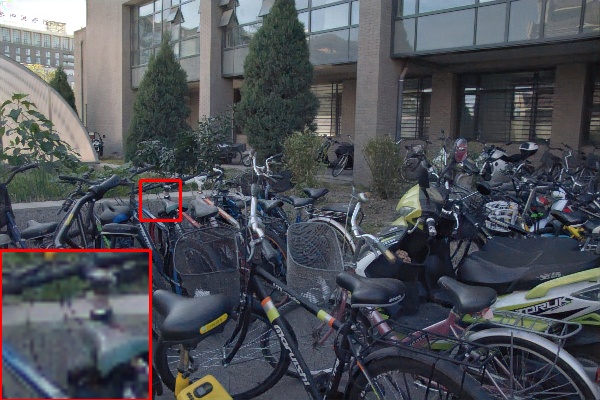} &\hspace{-4mm}
\includegraphics[width = 0.16\linewidth]{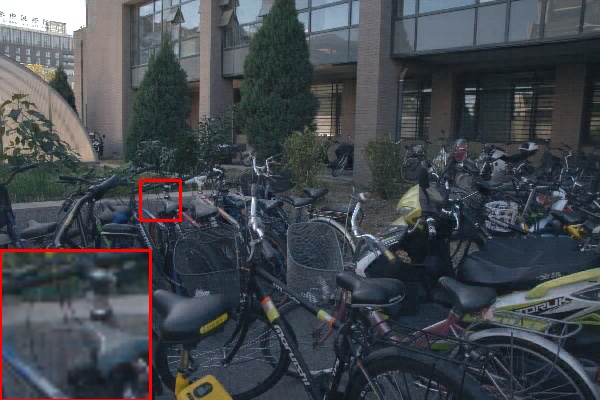} &\hspace{-4mm}
\includegraphics[width = 0.16\linewidth]{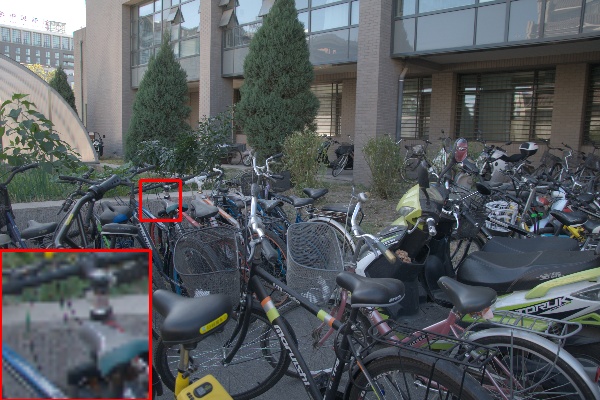} &\hspace{-4mm}
\\[0.2mm]

\hspace{-4mm}
\includegraphics[width = 0.16\linewidth]{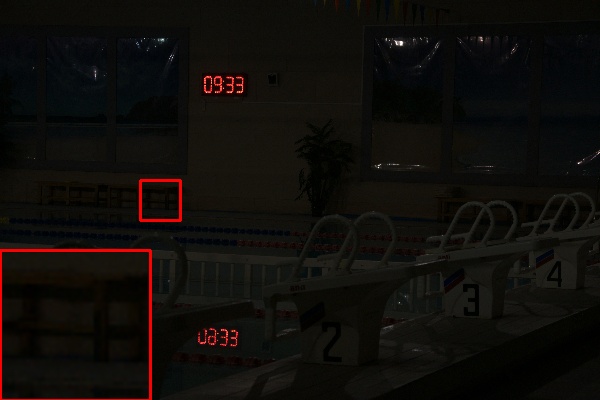} &\hspace{-4mm}
\includegraphics[width = 0.16\linewidth]{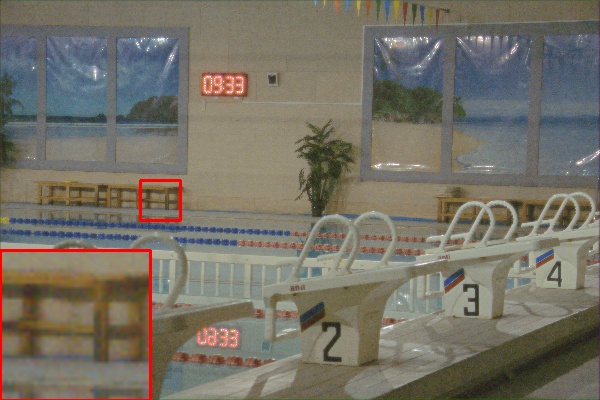} &\hspace{-4mm}
\includegraphics[width = 0.16\linewidth]{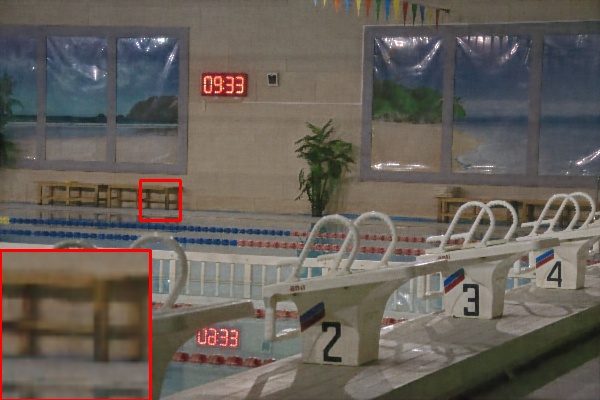} &\hspace{-4mm}
\includegraphics[width = 0.16\linewidth]{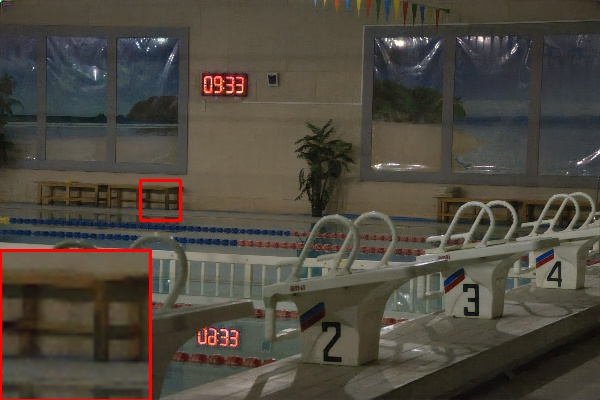} &\hspace{-4mm}
\includegraphics[width = 0.16\linewidth]{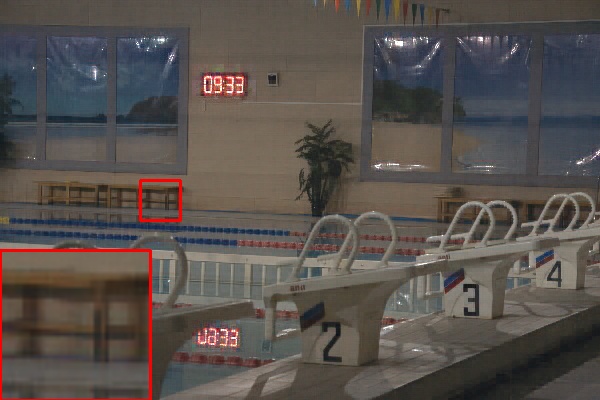} &\hspace{-4mm}
\includegraphics[width = 0.16\linewidth]{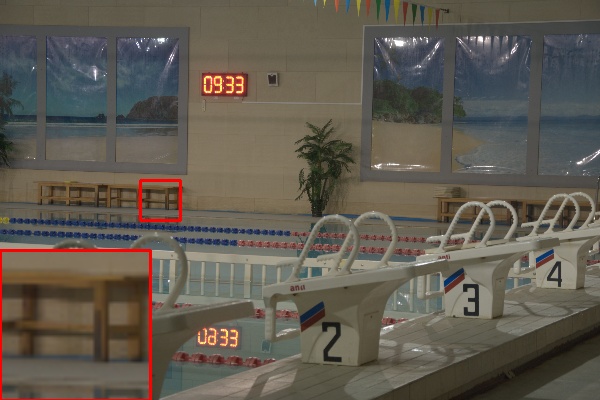} &\hspace{-4mm}
\\[0.2mm]

Input &\hspace{-4mm}   IAT  &\hspace{-4mm}   Retinexformer  &\hspace{-4mm} FourLLIE &\hspace{-4mm} Ours &\hspace{-4mm} GT
\\
\end{tabular}
\end{center}
\caption{Visual comparison of different methods on the LOLv1 \cite{Chen2018Retinex} and LOLv2-real \cite{yang2021sparse} datasets. The first two lines are the result of the LOLv1 dataset and the last two lines are the results of the LOLv2-real dataset. It is obvious that our model has more natural enhancement results and brings a pleasant visual experience.}\label{Visual_comparison}
\end{figure*}

\noindent{\textbf{Implementation Details}}. We implemented SDTL through Pytorch \cite{paszke2019pytorch}, and this model was trained 1000 epochs on 8 NVIDIA V100-SXM2-32GB. We achieve multi-gpu parallel training through the accelerate framework \footnote{\url{https://github.com/huggingface/accelerate}}. The optimizer adopts Adam \cite{kingma2014adam}, the initial learning rate is set to 5e-4 and batch size is set to 8. We use StepLR to gradually adjust the learning rate during the training process, where the step size is set to 50 and gamma is set to 0.90. In addition, we randomly crop 256 patches as training samples in low-light/normal-light image pairs during training. It should be noted that we do not use any data augmentation and training strategies except for cropping, which is an end-to-end model. We use 200 diffusion steps for training and in the inference stage, we adopt the DDIM \cite{song2020denoising} and set 10 steps for image enhancement. In night segmentation, we use MMSegmentation \cite{mmseg2020}
for experiments. 


\begin{figure*}[t]
\begin{center}
\renewcommand\arraystretch{0.1} 
\begin{tabular}{cccccc}
\hspace{-4mm}
\includegraphics[width = 0.19\linewidth]{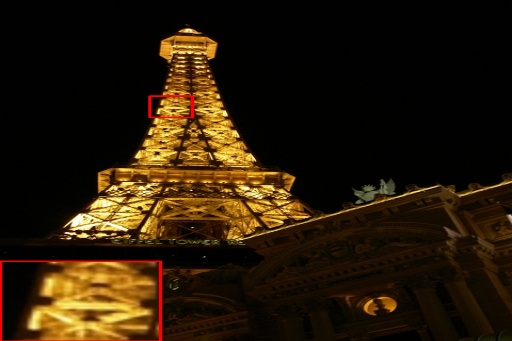} 
&\hspace{-4mm}
\includegraphics[width = 0.19\linewidth]{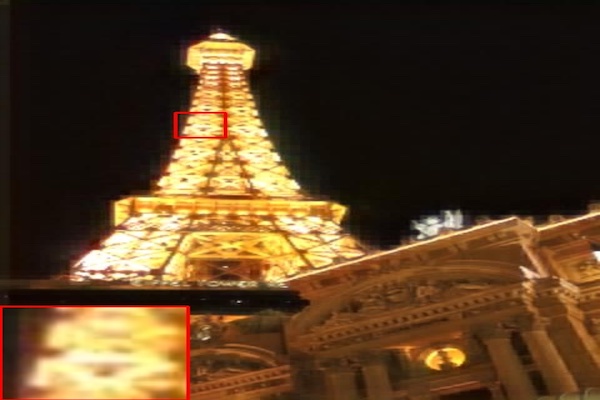} 
&\hspace{-4mm}
\includegraphics[width = 0.19\linewidth]{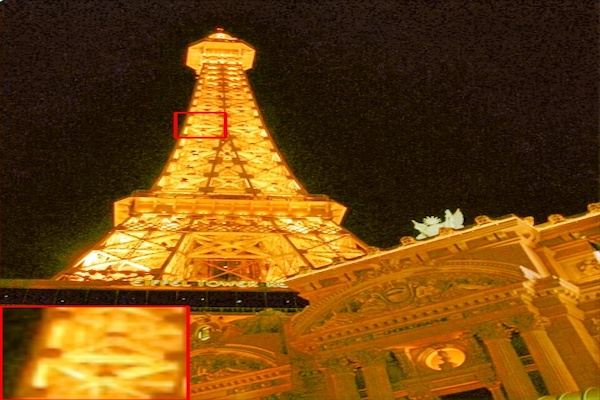} &\hspace{-4mm}
\includegraphics[width = 0.19\linewidth]{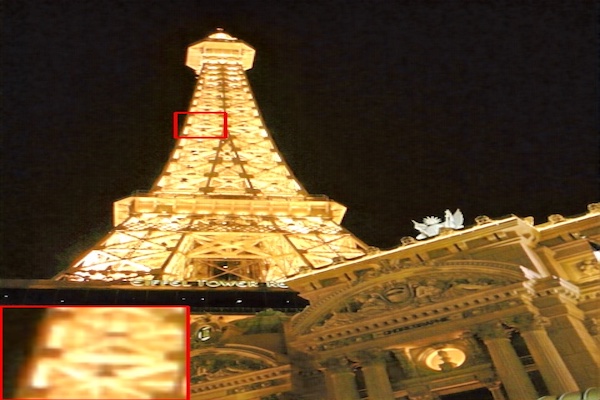} &\hspace{-4mm}
\includegraphics[width = 0.19\linewidth]{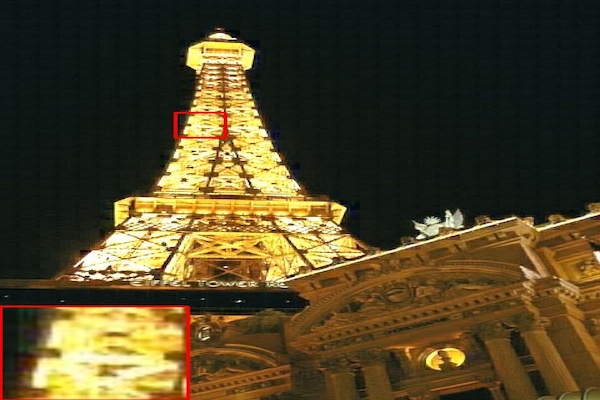} 
&\hspace{-4mm}

\\
\hspace{-4mm}
\includegraphics[width = 0.19\linewidth]{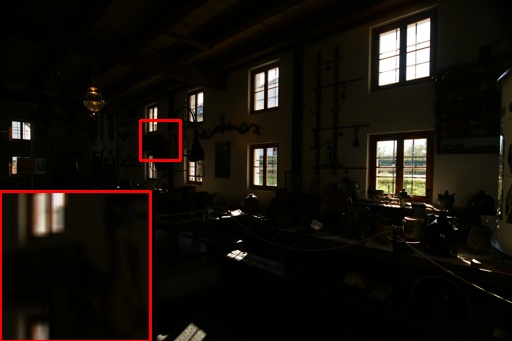} 
&\hspace{-4mm}
\includegraphics[width = 0.19\linewidth]{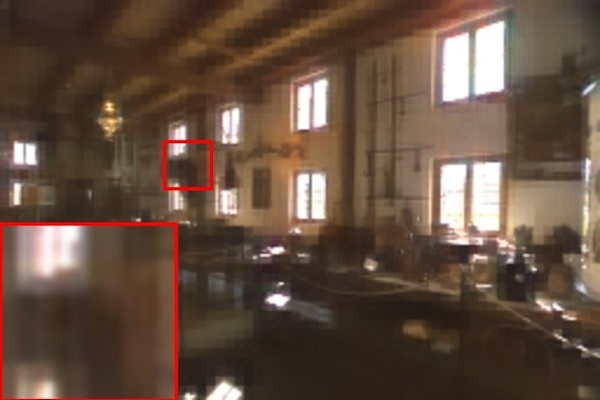} &\hspace{-4mm}
\includegraphics[width = 0.19\linewidth]{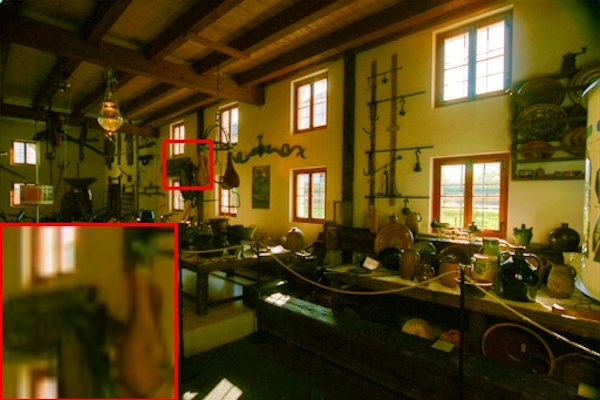} &\hspace{-4mm}
\includegraphics[width = 0.19\linewidth]{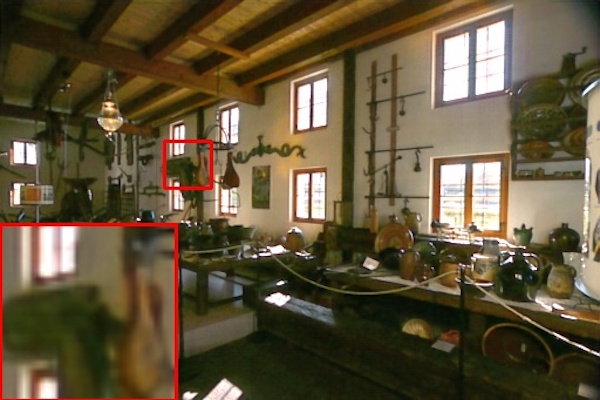} &\hspace{-4mm}
\includegraphics[width = 0.19\linewidth]{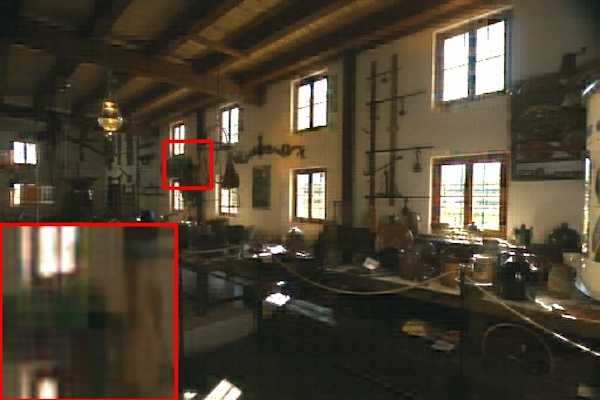} &\hspace{-4mm}

\\[0.2mm]
Input &\hspace{-4mm} DiffLL &\hspace{-4mm} FourLLIE  &\hspace{-4mm} LLFormer &\hspace{-4mm} Ours 
\\
\end{tabular}
\caption{Comparison on DICM \cite{lee2013contrast} (row 1) and MEF \cite{ma2015perceptual} (row 2). SDTL still has good generalization in invisible real scenes. }\label{NIQE_vis}\end{center}
\end{figure*}
\subsection{Comparison with SOTA Methods}

\subsubsection{Quantitative Results}

Table \ref{comparisons-main} demonstrated the comparison on the LOLv1 \cite{Chen2018Retinex}, LOLv2-real \cite{yang2021sparse} and LSRW \cite{hai2023r2rnet} datasets. Obviously, our SDTL achieved advanced performance in three datasets in all the comparative SOTA methods. {Specifically, compared with the second best model DiffLL, our method has been significantly improved in most of the metrics of most datasets. Our SDTL increased by 1.95dB, 0.024 and 2.725 respectively in PSNR, SSIM, and FID compared to DiffLL on the LSRW dataset. On the LOLv1 dataset, our method has achieved the best results on PSNR (27.34), SSIM (0.862) and LPIPS (0.118) and the second-best performance on FID (50.432). On the LOLv2 dataset, our method obtained the best results on PSNR (28.85) and FID (43.237). Due to the domain gap between the training set and the test set, many methods did not have stable performance. Our method has achieved the best or the second-best results in LPIPS and FID on the three datasets, which shows that our model produced satisfied visual quality.} 
We also tested NIQE on five unpaired datasets DICM, LIME, MEF, MEF, NPE, and VV, as shown in Table \ref{NIQE_Compa}. The lower the values of NIQE, the better the visual quality of the enhanced image. We achieved the best results in LIME and MEF datasets and the second best results in VV and NPE datasets. We achieved the average best result of the five datasets, with an average NIQE of 3.565. 
This demonstrated that our SDTL has fully explored the potential of DiT and has better robustness in invisible real scenes.

\subsubsection{Qualitative Results}

The visualization of LOLv1 and LOLv2-real datasets is shown in Fig. \ref{Visual_comparison}. We enlarged part of the local area to the bottom left of each image. The first two lines in the figure are the results of the LOLv1 dataset and the last two lines are the results of the LOLv2-real dataset, our SDTL showed pleasant perceptual quality. {Specifically, IAT \cite{cui2022illumination} and Retinexformer \cite{cai2023retinexformer} generated overly smooth textures and rich details and the exposure is unstable. Our method effectively improved the contrast, restored clearer details and suppressed noise. Retinexformer \cite{cai2023retinexformer} and FourLLIE \cite{wang2023fourllie} had different degrees of color distortion or noise amplification.} In contrast, our model can effectively enhance areas with low contrast and steadily restore color without introducing noise. In addition, we also compared the comparison between DICM and MEF of unpaired datasets, as shown in Fig. \ref{NIQE_vis}. LLFormer \cite{jie2023llformer} and DiffLL \cite{jiang2023low} did not restore details and colors well and the texture is not rich. Obviously, our method is more natural in color, and the texture is more delicate than other models. This shows that the structure guide in DiT is effective and also performs well in quantitative evaluation.

\begin{figure*}[t]
\begin{center}
\renewcommand\arraystretch{0.1} 
\begin{tabular}{cccccc}
\hspace{-4mm}
\includegraphics[width = 0.19\linewidth]{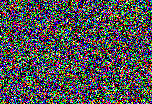} 
&\hspace{-4mm}
\includegraphics[width = 0.19\linewidth]{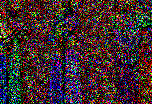} 
&\hspace{-4mm}
\includegraphics[width = 0.19\linewidth]{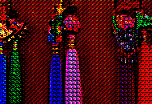} &\hspace{-4mm}
\includegraphics[width = 0.19\linewidth]{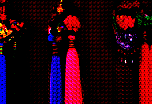} &\hspace{-4mm}
\includegraphics[width = 0.19\linewidth]{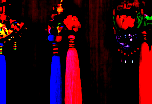} 
&\hspace{-4mm}

\\
\hspace{-4mm}
\includegraphics[width = 0.19\linewidth]{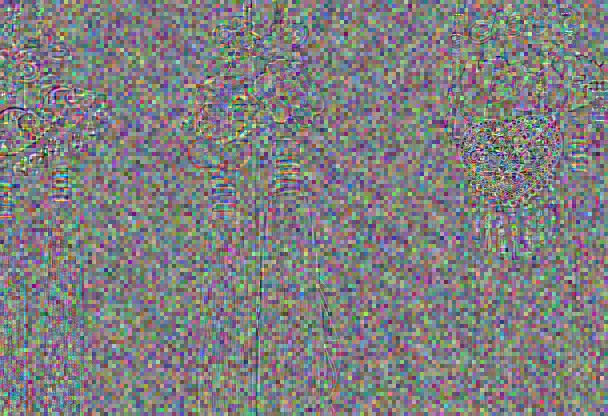} 
&\hspace{-4mm}
\includegraphics[width = 0.19\linewidth]{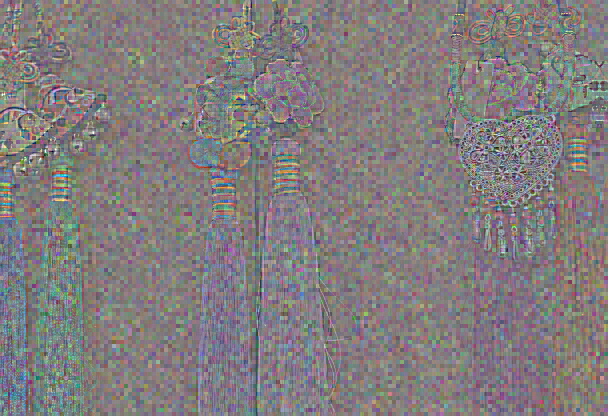} &\hspace{-4mm}
\includegraphics[width = 0.19\linewidth]{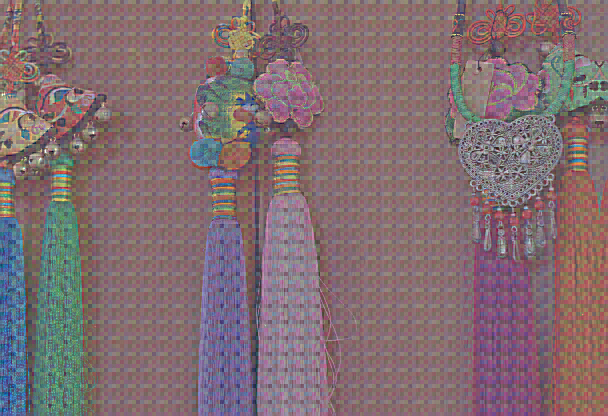} &\hspace{-4mm}
\includegraphics[width = 0.19\linewidth]{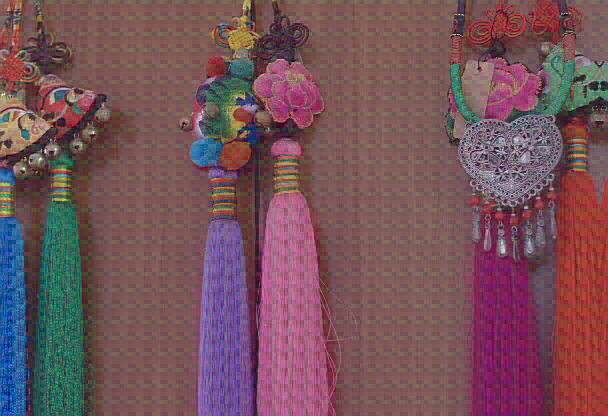} 
&\hspace{-4mm}
\includegraphics[width = 0.19\linewidth]{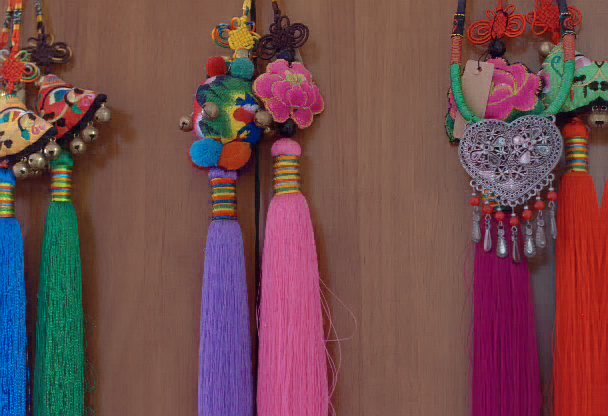} &\hspace{-4mm}
\\[0.2mm]
Step 0 (Input) &\hspace{-4mm} Step 1 &\hspace{-4mm} Step 3  &\hspace{-4mm} Step 6 &\hspace{-4mm} Step 10 (Output)
\end{tabular}
\caption{{Visualization of reverse denoising processing in SDTL. The first line represents the frequency information in the denoising step, and the second line represents the corresponding enhanced image. This process shows the powerful ability of the diffusion model in low-light enhancement and gradually removes the noise and artifacts in low-light images. }}\label{Denoising_vis}\end{center}
\end{figure*}

\noindent{\textbf{Visualization of Reverse Denoising processing in SDTL: }}{We visualize the process of SDTL under low light enhancement, which is the reverse denoising process of the diffusion model, as shown in Fig. \ref {Denoising_vis}. We use 10 sampling time steps in the process of reverse denoising and show the results for step 0, step 1, step 3, step 6, and step 10, respectively. The first line is the frequency information after wavelet transformation in a certain step, and the second line is the corresponding enhanced results. According to visualization, it can be observed that our SDTL gradually generates normal-light images during the reverse denoising process, and removes noise and artifacts. The iterative refinement process of the diffusion model is the benefit of detail recovery.}

\begin{figure*}[t]
\begin{center}
\renewcommand\arraystretch{0.1} 
\begin{tabular}{ccccccc}
\hspace{-4mm}
\includegraphics[width = 0.16\linewidth]{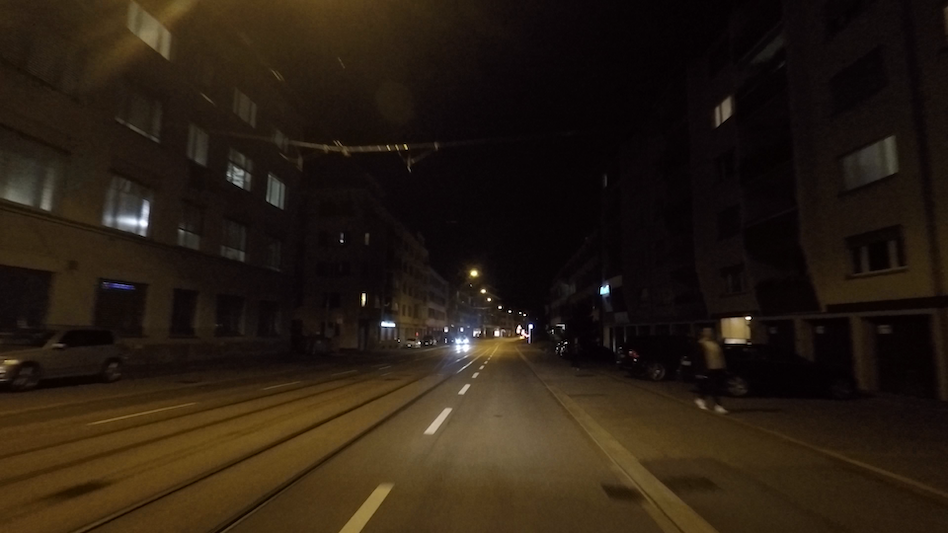} 
&\hspace{-4mm}
\includegraphics[width = 0.16\linewidth]{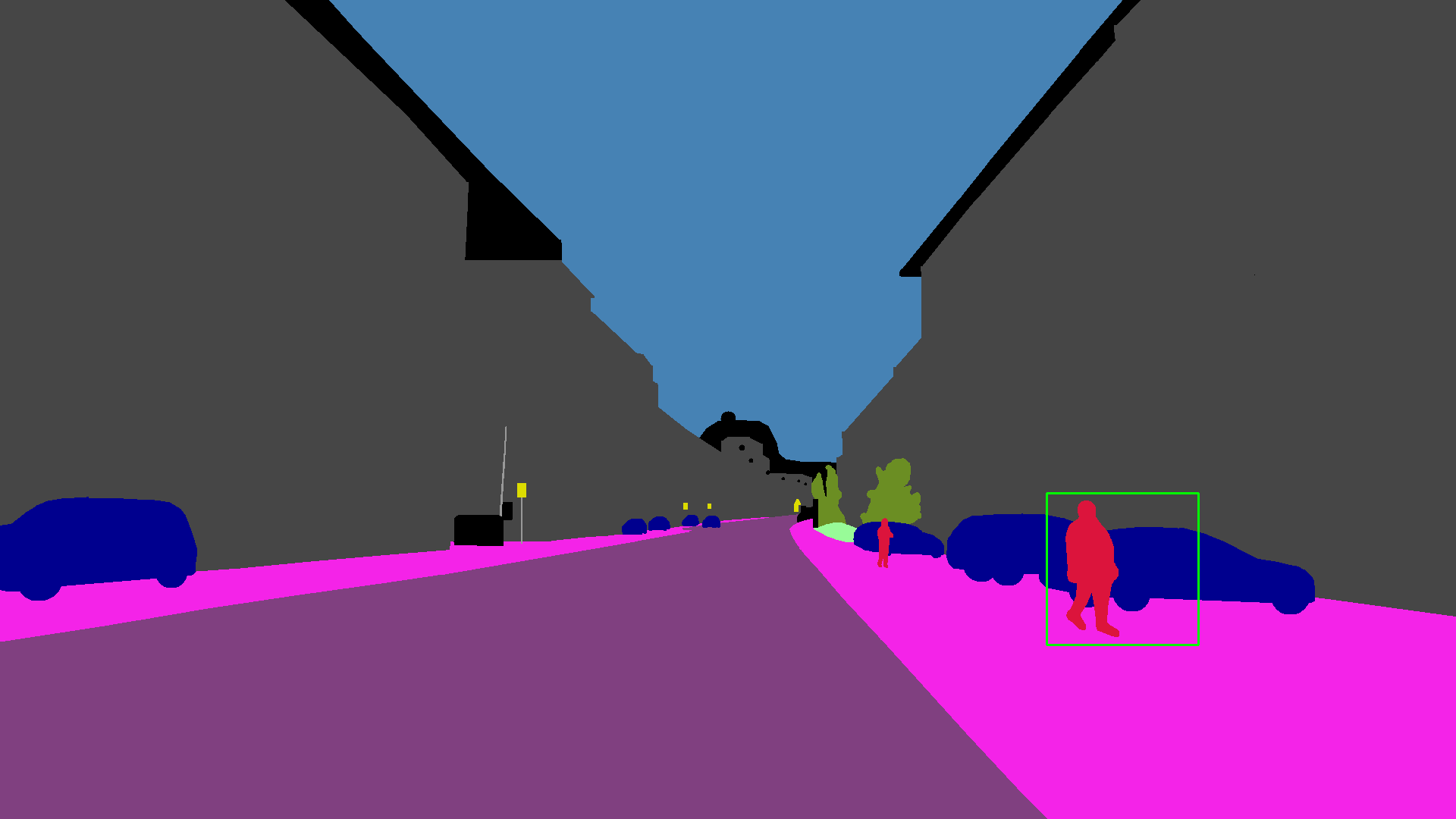} 
&\hspace{-4mm}
\includegraphics[width = 0.16\linewidth]{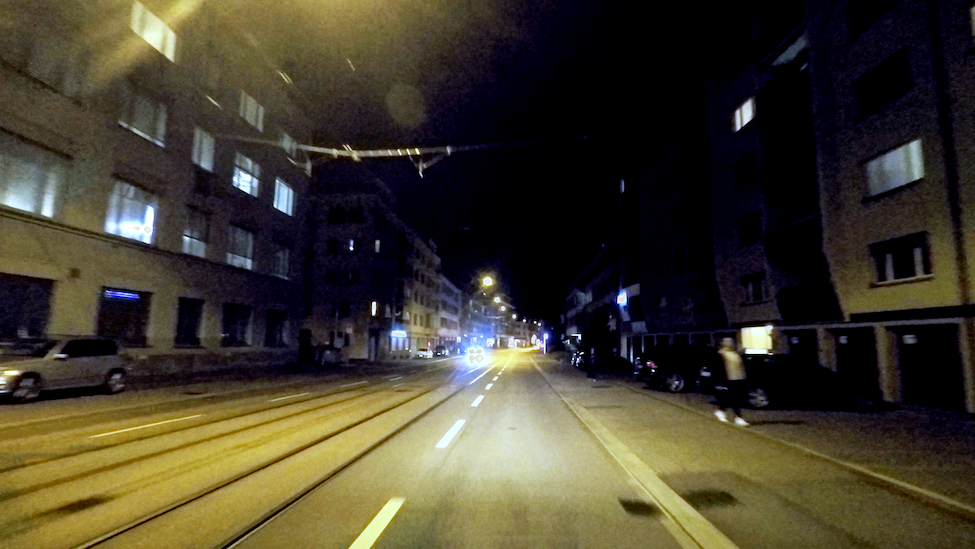} &\hspace{-4mm}
\includegraphics[width = 0.16\linewidth]{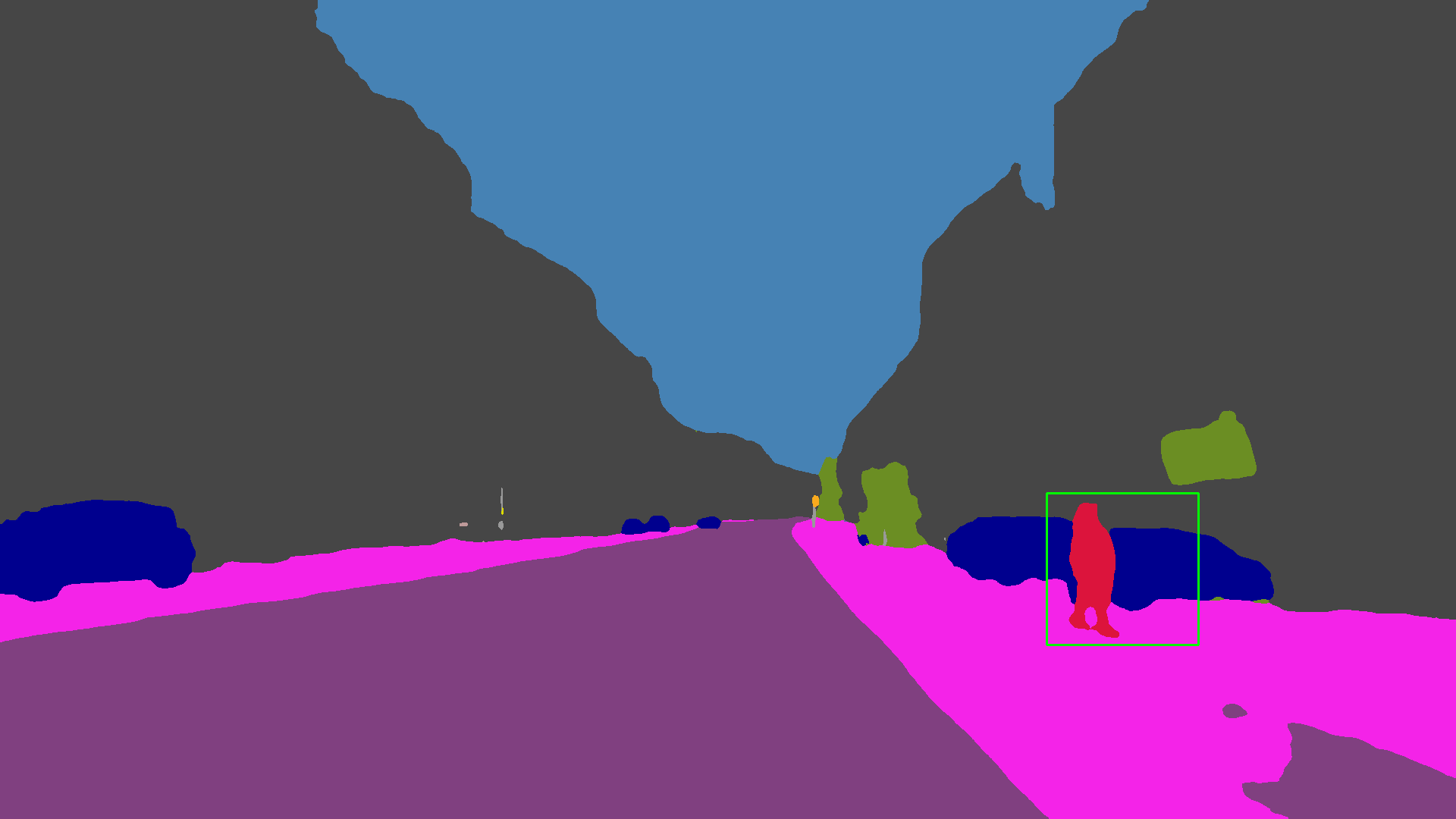} &\hspace{-4mm}
\includegraphics[width = 0.16\linewidth]{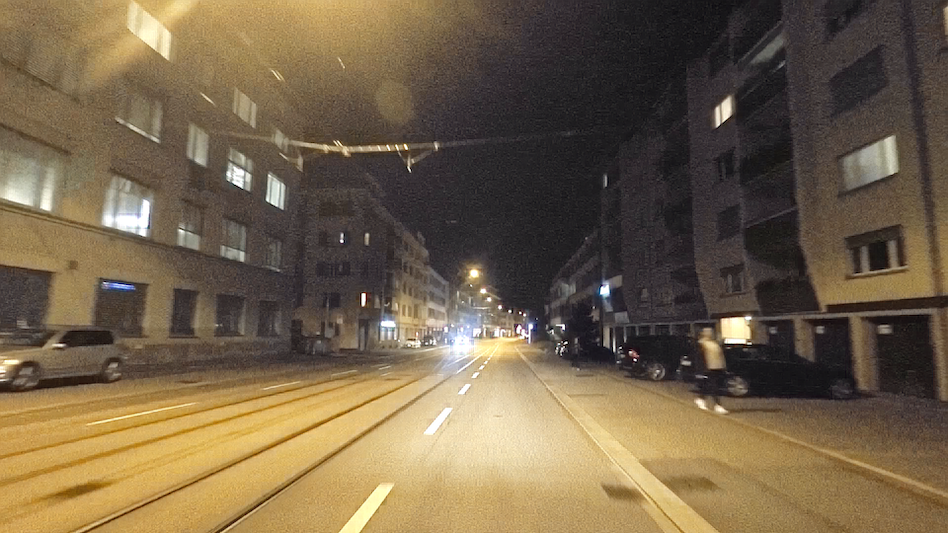} 
&\hspace{-4mm} 
\includegraphics[width = 0.16\linewidth]{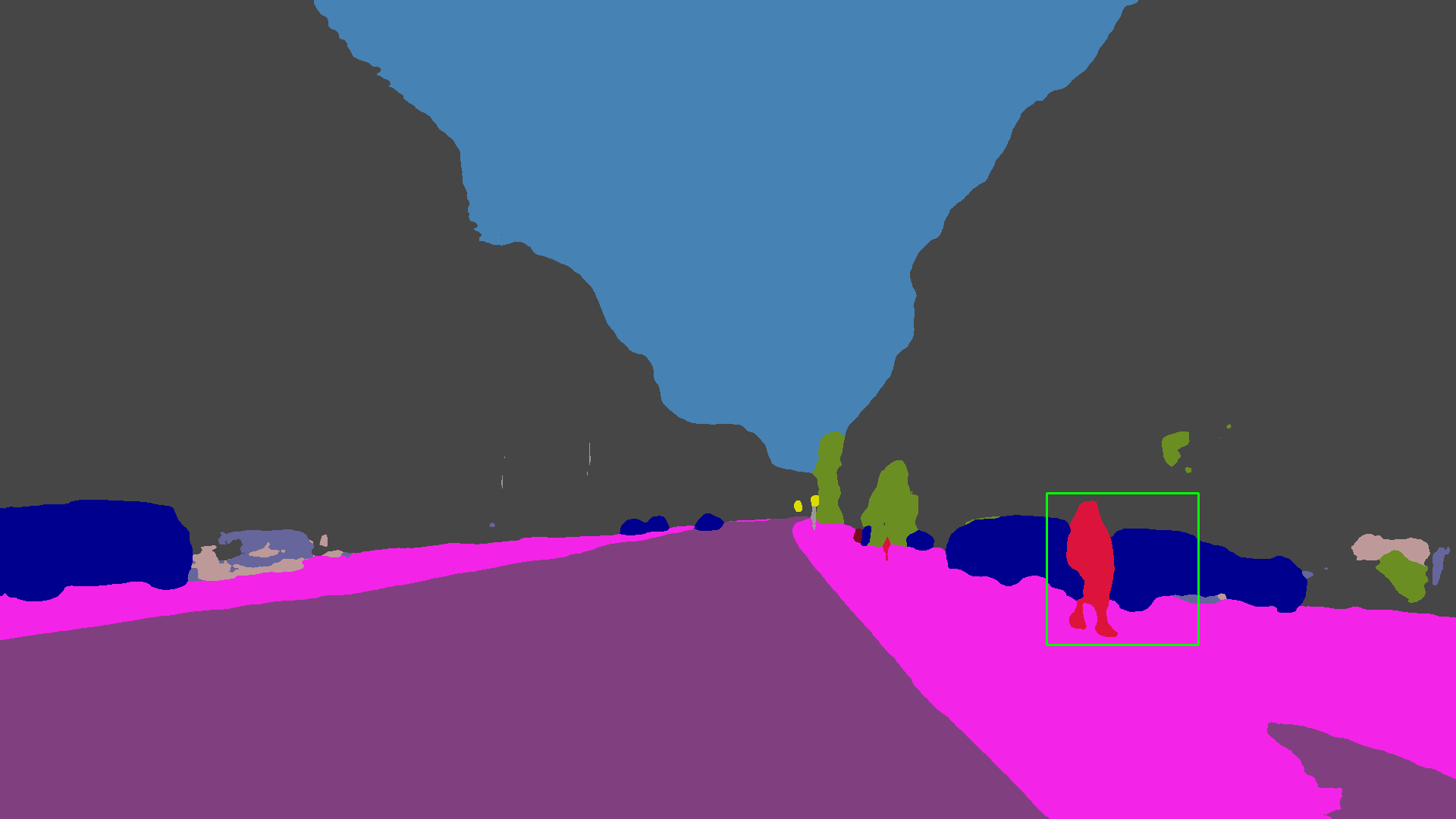} 
&\hspace{-4mm}
\\[0.2mm]
Origin &\hspace{-4mm} Label &\hspace{-4mm} DeepLPF Enhance &\hspace{-4mm} DeepLPF Seg &\hspace{-4mm} SDTL Enhance &\hspace{-4mm} SDTL Seg 
\end{tabular}
\caption{Visualization of night segmentation on the ACDC \cite{sakaridis2021acdc} dataset. We visualized the enhanced image by the enhancement model and the corresponding predictive segmentation mask.}\label{Night_vis}
\end{center}
\end{figure*}


\subsubsection{Application of Night Segmentation}
We adopted DeepLabv3+ \cite{chen2018encoder} as the segmentor, using SDTL to enhance the images of the ACDC dataset for the training of night segmentation. {We only use the part of the night scene on the ACDC dataset, with about 1,000 images and corresponding high-quality pixel-level labels. We only use the part of the ACDC night scene, which contains about 1,000 images. We use the training set to learn the model and the valid set to evaluate the model. To compare the models fairly, we set the same hyperparameters and training strategies. The training input is the image obtained from the enhanced model, and the label does not change. The experimental results are shown in Table \ref{Segmentation}, and our SDTL has greatly improved the performance of the segmentation model. Our method is 0.76\% and 0.4\% higher than DiffLL in mIoU and Fscore respectively. Compared with the baseline, our model has improved by 1.77 and 2.68 on Fscore (74.18\%) and mIoU (62.25\%) respectively. In addition, we visualize the enhanced ACDC and segmentation results, as shown in Fig. \ref{Night_vis}. From the first column of the figure, we compare the enhanced image effect. The contrast ratio of DeepLPF is poor and the far area is dark. The increment and denoising of our model at the distance of the street view are significantly better than that of DeepLPF. In the comparison of the second column, ours is more accurate on some objects in segmentation results. }

\subsection{Ablation Studies}
\noindent{\textbf{Effect of the number of SDT block: }}
The impact of the depth (\textit{i.e.,} the SDT block number) of Denoising DiT has been investigated on the LOLv2-real dataset, which is shown in Table \ref{depth}. The results demonstrated that the larger the depth, the stronger the learning ability of the model, it will also bring more parameters. Our model still performed well even with a small number of parameters (depth=2). This demonstrated that we can adjust the depth of the network in time and adapt to different computing environments as needed.

\begin{table}[t]
\caption{Effect of the depth (\textit{i.e.,} the number of SDT block) of Denoising DiT on the LOLv2-real dataset. 
 }\label{depth}
\renewcommand{\arraystretch}{1.3} 
\centering
\begin{tabular}{c|cccc}
\hline
\textbf{Depth}  & \textbf{PSNR}$\uparrow$ & \textbf{SSIM}$\uparrow$ & \textbf{LPIPS}$\downarrow$   &\textbf{Parameters} (M) \\
\hline
2	&26.67	&0.822	&0.215	 & 11.08 \\
4	&27.66	&0.866	&0.148	 & 18.77 \\
6	&28.85	&0.875	&0.131	 & 24.84 \\
\hline
\end{tabular}
\end{table}

\begin{table}[t]
\caption{Ablation studies on the LOLv2-real dataset. 
The results show the effectiveness of each component in our model. }\label{Ablation}
\renewcommand{\arraystretch}{1.3} 
\centering
\begin{tabular}{l|ccc}
\hline
Methods  & \textbf{PSNR}$\uparrow$ & \textbf{SSIM}$\uparrow$ & \textbf{LPIPS}$\downarrow$  \\
\hline
w/o SEM-enhance &  26.89  &  0.819 &   0.212 	\\
w/o SEM-fusion &  26.40  &  0.852 &   0.160 	\\
w/o SEM &  25.92  &  0.763 &   0.291 	\\
w/o DiT &  26.65  &  0.842  &  0.196     \\
w/o SAB &  27.41  &  0.852 &   0.160  \\
w ALL   &  28.85  &  0.875 &0.131	\\
\hline
\end{tabular}
\vspace{-0.2cm}
\end{table}

\noindent{\textbf{Effect of each component in our SDTL: }}
We carry out the following studies to investigate the effectiveness of the key modules in SDTL by deleting different components separately:
\begin{itemize}
    \item "w/o SEM-enhance": Only remove the enhancement stage in SEM;
    \item "w/o SEM-fusion": Only remove the fusion stage in SEM;
    \item "w/o SEM": Remove all parts of SEM, including enhancement and fusion;
    \item  "w/o DiT": The structure of DiT is not adopted and replaced with the original DDPM;
    \item  "w/o SAB": Remove SAB;
    \item "w ALL": Keep all components (SDTL)
\end{itemize}
We conducted experiments on ablation studies on the LOLv2-real dataset, and the results are shown in Table \ref{Ablation}. The comparison of "w/o SEM-enhance" and "w/o SEM" illustrates the necessity of complementary fusion of different frequency bands in the design of SEM, especially in SSIM and LPIPS, which have improved by 0.056 and 0.052 respectively. Through the comparison of "w/o SEM-fusion" and "w/o SEM", we found that SSIM and LPIPS increased by 0.089 and 0.131 respectively, it shows that the structural prior obviously enhances the texture of high-frequency information, which is beneficial to the recovery of information in the dark environment. Comparing "w/o DiT" and "w/o SAB" with "w ALL" highlights the advantages of the Transformer structure in the diffusion model, which not only delivers improved performance but also offers powerful flexibility and scalability due to the customizable number of blocks, embedding dimensions, and the number of self-attention heads in DiT. This allows for easy adjustments to the complexity of the model, making it more versatile than DDPM. From "w/o SAB", our method was improved by 0.023 and 0.029 respectively in SSIM and LPIPS. This result demonstrated that SAB effectively pays more attention to the texture-rich tokens while reducing the noise area. All in all, our framework brought superior improvement for low-light enhancement.

\begin{table}[t]
\caption{{Effect of the number of SEM on the LOLv2-real dataset.} }\label{SEM_number}
\renewcommand{\arraystretch}{1.3} 
\centering
\begin{tabular}{cc|ccc}
\hline
\textbf{SEM Number} && \textbf{PSNR}$\uparrow$ & \textbf{SSIM}$\uparrow$ & \textbf{LPIPS}$\downarrow$  \\
\hline
0	&&25.92	&0.763	&0.291 \\
1	&&26.79	&0.813	&0.237 \\
2	&&28.85	&0.875	&0.131 \\
\hline
\end{tabular}
\end{table}

\begin{table}[t]
\caption{Analysis of the structure guidance in SEM on the LOLv2-real dataset. }\label{stru_guidance}
\renewcommand{\arraystretch}{1.3} 
\centering
\begin{tabular}{cc|ccc}
\hline
\textbf{Method}  && \textbf{PSNR}$\uparrow$ & \textbf{SSIM}$\uparrow$ & \textbf{LPIPS}$\downarrow$ \\
\hline
w/o SEM-enhance &&  26.89  &0.819 &0.212 	\\
Self-Guidance	&&27.06	&0.833	&0.189	 \\
Structure-Guidance	&&28.85	&0.875	&0.131	 \\
\hline
\end{tabular}
\end{table}

\begin{table}[t]
\caption{Effect of the patch size of Denoising DiT on the LOLv2-real dataset. }\label{patchsize}
\renewcommand{\arraystretch}{1.3} 
\centering
\begin{tabular}{cc|ccc}
\hline
\textbf{Patch Size}  && \textbf{PSNR}$\uparrow$ & \textbf{SSIM}$\uparrow$ & \textbf{LPIPS}$\downarrow$ \\
\hline
2	&&\multicolumn{3}{c}{Out of Memory}	 \\
4	&&28.85	&0.875	&0.131	 \\
8	&&26.52	&0.824	&0.228	 \\
\hline
\end{tabular}
\end{table}

\begin{table}[t]
\caption{Effect of the embedding size of Denoising DiT on the LOLv2-real dataset. }\label{embeddingsize}
\renewcommand{\arraystretch}{1.3} 
\centering
\begin{tabular}{cc|ccc}
\hline
\textbf{Embedding Size}  && \textbf{PSNR}$\uparrow$ & \textbf{SSIM}$\uparrow$ & \textbf{LPIPS}$\downarrow$ \\
\hline
192	&&27.39	&0.863	&0.162	 \\
384	&&28.85	&0.875	&0.131	 \\
768	&&27.77	&0.863	&0.154	 \\
\hline
\end{tabular}
\end{table}

\noindent{\textbf{Effect of the SEM module: }}We analyzed the impact of different SEM numbers on the LOLv2 dataset, as shown in Table \ref{SEM_number}. The results showed that two SEMs have a better effect than one, which shows that the use of SEM in two wavelet transforms gradually restored the details of the frequency and achieved better results in the subsequent denoising network. Meanwhile, two wavelet transforms reduced the amount of calculation again. Meanwhile, we analyzed the impact of structure guidance in SEM, as shown in Table~\ref{stru_guidance}. Quantitative results demonstrated that structure guidance was necessary and achieved superior performance compared to self-guidance. Structure guidance adopts the details of the edge map, guiding the model to recover the high-frequency information from low-light images.

\noindent{\textbf{Effect of the patch size and embedding size of Denoising DiT:} We constructed the analysis of patch size in the SDT block on the LOLv2-real dataset, as shown in Table~\ref{patchsize}. It demonstrated that the smaller the patch size, the superior the network granularity, meanwhile the computational burden will also increase. Specially, the high patch size will be out of memory on the GPUs. Therefore, increasing the patch size within a range achieves better performance of our model. In addition, we also analyzed the embedding size of SDT block, as shown in Table~\ref{embeddingsize}. This demonstrated that the best embedding size is 384. Although the performance of our model from 192 to 384 has improved, the continuous improvement resulted in a decline of effect. We need to set the suitable embedding size to get stable model performance. }

\section{Conclusion}

In this paper, we introduce firstly Diffusion Transformer (DiT) into low-light enhancement and design a framework named Structure-guided Diffusion Transformer for Low-light (SDTL). We design Structure Enhancement Module (SEM) and Structure-guided Attention Block (SAB) respectively to improve network performance by using structure prior. The experimental results show that our method achieves advanced performance on 8 benchmark datasets and improves the effect of the segmentation in the night dataset. In the future, we will extend our idea to other tasks for 3D engineering applications \cite{li20233d,li2023tpnet}.






\ifCLASSOPTIONcaptionsoff
  \newpage
\fi



\bibliographystyle{IEEEtran}
\bibliography{reference}

%




\end{document}